\documentclass[shortpaper]{clv2025}

\jvol{vv}
\jnum{nn}
\jyear{2025}

\dochead{Short Paper} 

\pageonefooter{Action editor: Minlie Huang. Submission received: 3 March 2025; revised version received: 30 July 2025; accepted for publication: 24 October 2025.}

\runningtitle{Vocabulary Expansion of LLMs with 0.01GB of Target Language Text}
\runningauthor{Yamaguchi et al.}

\usepackage[table]{xcolor}
\usepackage{amsmath}
\usepackage{amssymb}
\usepackage{amsfonts}
\usepackage{graphicx}
\usepackage{xurl}
\usepackage{booktabs}
\usepackage{multirow} 
\usepackage{url}
\usepackage{here}
\usepackage{tabularx}
\usepackage{xcolor}
\usepackage{hhline}
\usepackage{hyperref}
\definecolor{darkblue}{rgb}{0, 0, 0.5}
\hypersetup{colorlinks=true,citecolor=darkblue, linkcolor=darkblue, urlcolor=darkblue}

\begin{document}

\title{How Can We Effectively Expand the Vocabulary of LLMs with 0.01GB of Target Language Text?}

\author{Atsuki Yamaguchi$^{1}$\thanks{Regent Court, 211 Portobello, Sheffield, S1 4DP, United Kingdom. E-mail: \texttt{ayamaguchi1@sheffield.ac.uk}.}, Aline Villavicencio$^{1,2,3,4}$, Nikolaos Aletras$^{1}$}

\affilblock{
    \affil{University of Sheffield, United Kingdom}
    \affil{University of Exeter, United Kingdom}
    \affil{The Alan Turing Institute, United Kingdom}
    \affil{Federal University of Rio Grande do Norte, Brazil}
}

\maketitle

\begin{abstract}
Large language models (LLMs) have shown remarkable capabilities in many languages beyond English.
Yet, LLMs require more inference steps when generating non-English text due to their reliance on English-centric tokenizers and vocabulary, resulting in higher usage costs to non-English speakers.
Vocabulary expansion with target language tokens is a widely used cross-lingual vocabulary adaptation approach to remedy this issue.
Despite its effectiveness in inference speedup, previous work on vocabulary expansion has focused on high-resource settings assuming access to a substantial amount of target language data to effectively initialize the embeddings of the new tokens and adapt the LLM to the target language.
However, vocabulary expansion in low-resource settings has yet to be explored.
In this article, we investigate vocabulary expansion in low-resource settings by considering embedding initialization methods and continual pre-training strategies.
Through extensive experiments across typologically diverse languages, tasks and models, we establish a set of strategies to perform vocabulary expansion for faster inference, while striving to maintain competitive downstream performance to baselines.
This is achieved with only 30K sentences ($\sim$0.01GB text data) from the target language.\footnote{Our code and models are available via \href{https://github.com/gucci-j/lowres-cve}{GitHub}.}
\end{abstract}

\section{Introduction}

\begin{figure}[!t]
\begin{center}
\includegraphics[width=0.9\columnwidth]{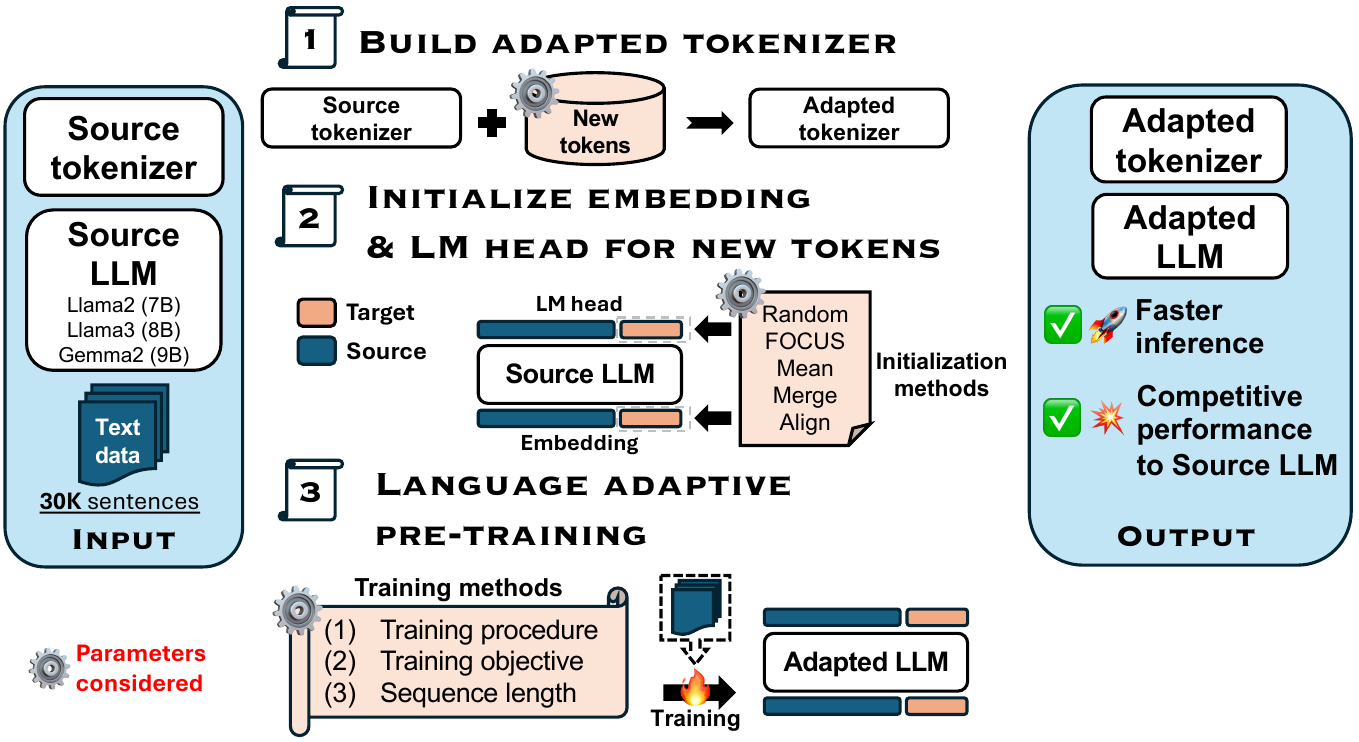}
\caption{
We address the challenge of effectively expanding vocabulary for LLMs in low-resource settings.
This is crucial for reducing inference steps when generating non-English text, as LLMs often rely on English-centric tokenizers and vocabulary.
Our approach explores various adaptation strategies (\includegraphics[width=1em]{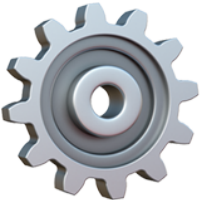}) to achieve inference speedups while aiming to retain competitive performance.
Our recommended strategy combines heuristic-based parameter initialization for new tokens with fine-tuning the top and bottom two layers of the model, using a short input sequence length and a multi-token prediction objective~\cite{pmlr-v235-gloeckle24a}.
}
\label{fig:overview}
\end{center}
\end{figure}

Large language models (LLMs) have strong capabilities in English and other languages~\cite{Achiam2023GPT4TR,openai2024gpt4ocard,Touvron2023Llama2O,Jiang2023Mistral7,groeneveld-etal-2024-olmo,deepseekai2025deepseekr1incentivizingreasoningcapability}.
Yet, processing non-English texts with LLMs is challenging. They suffer from tokenization overfragmentation (see Figure \ref{fig:flores} for quantitative analysis) and thus require more inference steps due to the reliance on English-centric tokenizers and vocabulary, resulting in higher utility costs for non-English speakers~\cite{ahia-etal-2023-languages,petrov2023language,ali-etal-2024-tokenizer}. 

Cross-lingual vocabulary adaptation (CVA) via vocabulary expansion has been proposed to adapt LLMs (including their tokenizers) to specific target languages~\cite[\textit{inter alia}]{Cui2023EfficientAE,fujii2024continual,choi-etal-2024-optimizing,tejaswi-etal-2024-exploring,mundra-etal-2024-empirical}. Vocabulary expansion approaches extend the vocabulary of a source model with tokens from a target language, followed by continual pre-training on target language data. A wide range of language-specific LLMs derived from an English-centric LLM such as Llama2~\cite{Touvron2023Llama2O} have been made available following this approach, including Chinese~\cite{Cui2023EfficientAE}, Tamil~\cite{Balachandran2023TamilLlamaAN}, Portuguese~\cite{Larcher2023CabritaCT}, and Japanese~\cite{fujii2024continual} models, \textit{inter alia}.
Vocabulary expansion improves inference speed but often assumes access to a substantial amount of target language data for adaptation.
For example, Chinese and Tamil Llamas make use of 20 and 12GB of target language text, respectively~\cite{Cui2023EfficientAE,Balachandran2023TamilLlamaAN}.
Given the guaranteed large number of model updates in high-resource settings, the embeddings of new tokens are often randomly initialized, followed by the standard continual pre-training with a causal language modeling (\textsc{clm}) objective~\cite{Cui2023EfficientAE,Balachandran2023TamilLlamaAN,Larcher2023CabritaCT,choi-etal-2024-optimizing}. 
However, it is not clear how effective this approach to vocabulary expansion is in low-resource settings.

In this article, we seek to (i) \textit{answer if this widely used adaptation approach under high-resource settings is as effective in low-resource settings}; and (ii) \textit{identify the best possible vocabulary expansion strategies for language adaptation in low-resource settings, while striving to maintain similar performance to the source model with faster inference} {\rm (Figure \ref{fig:overview})}.
Our key contributions are as follows:
\begin{itemize}
    \item We present the first systematic study of vocabulary expansion-based adaptation of generative LLMs (i.e three English-centric models) in low-resource settings (i.e. assuming only 30K sentences, $\sim$0.01GB text data or up to approximately 5M tokens), on two generation tasks (i.e. machine translation and summarization) and two classification tasks (i.e. multiple-choice reading comprehension and general knowledge and reasoning) across ten typologically diverse languages.

    \item Our results show that the popular vocabulary expansion approach in high-resource settings (i.e. random initialization and fine-tuning the full model with a \textsc{clm} objective) is not always optimal in low-resource settings.

    \item We find that target parameter initialization approaches that use heuristics information from the source and target tokenizers are more effective. Furthermore, fine-tuning the top and bottom two layers of the LLM using a multi-token prediction objective~\cite{pmlr-v235-gloeckle24a} works better than fine-tuning the full model with \textsc{clm}. Finally, using a short input sequence length by splitting longer text into multiple sentences allows for a larger number of model updates, mitigating underfitting.

    \item To better understand holistic aspects of vocabulary expansion in low-resource settings, we conduct a range of analyses, including the extent of source (English) knowledge retention and a direct comparison with vocabulary replacement.
   
\end{itemize}

\section{Background}

\subsection{Text Overfragmentation}
LLMs tend to overfragment text in underrepresented languages~\cite{rust-etal-2021-good,muller-etal-2021-unseen}. 
Overfragmentation has significant implications for non-English speakers, including higher API costs~\cite{ahia-etal-2023-languages, petrov2023language}, slower inference~\cite{hofmann-etal-2022-embarrassingly,sun-etal-2023-multi,petrov2023language}, and lower downstream performance~\cite{bostrom-durrett-2020-byte,rust-etal-2021-good,10.1145/3578707,fujii-etal-2023-different,ali-etal-2024-tokenizer}.

\begin{figure}[t]
\begin{center}
\includegraphics[width=0.8\columnwidth]{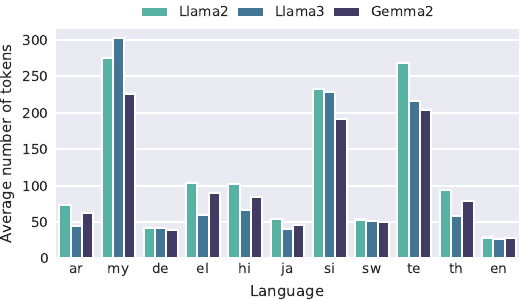}
\caption{
Average number of tokens on the FLORES-200 dev set across languages and models.
}
\label{fig:flores}
\end{center}
\end{figure}

To quantify this phenomenon, we calculate the average number of tokens on the FLORES-200~\cite{nllb-22} dev set across our target languages and models (Figure \ref{fig:flores}).
Following the established assumption that lower average token counts indicate more efficient tokenization~\cite{ahia-etal-2023-languages},\footnote{This assumption is valid specifically within the context of the FLORES-200 parallel machine translation corpus.} our analysis reveals two key observations: (1) Burmese, Sinhala, and Telugu demonstrate the most severe overfragmentation among our ten target languages, requiring at least 6.8x more tokens than English across all models, and (2) English consistently exhibits the most efficient tokenization.

\subsection{Cross-lingual Vocabulary Adaptation}
CVA methods have been proposed for adapting base LLMs to specific target languages for improving downstream performance and inference speed~\cite{Cui2023EfficientAE,Balachandran2023TamilLlamaAN,fujii2024continual,yamaguchi-etal-2024-empirical}. 
CVA via \textit{vocabulary expansion} incorporates new tokens into the source vocabulary~\cite{Balachandran2023TamilLlamaAN,Larcher2023CabritaCT,pipatanakul2023typhoonthailargelanguage,Lin2024MaLA500ML,Cui2023EfficientAE,kim2024efficienteffectivevocabularyexpansion,fujii2024continual,choi-etal-2024-optimizing,nguyen-etal-2024-seallms,tejaswi-etal-2024-exploring,mundra-etal-2024-empirical}.
CVA with \textit{vocabulary replacement} replaces the entire or partial source vocabulary with a new one from the target~\cite{ostendorff2023efficientlanguagemodeltraining,csaki2023efficientlyadaptingpretrainedlanguage,da-dalt-etal-2024-flor,remy2024transtokenization,yamaguchi-etal-2024-empirical,dobler2024language,cahyawijaya-etal-2024-cendol}.
More recent methods include a hypernetwork for tokenizer transfer~\cite{minixhofer2024zeroshot} and adapters for vocabulary alignment~\cite{han2024adaptersalteringllmvocabularies}.
CVA is typically followed by continual pre-training on target language data, often called language adaptive pre-training -- LAPT~\cite{chau-etal-2020-parsing}.

Vocabulary expansion is widely used for developing language-specific generative LLMs from a source model, e.g. Chinese~\cite{Cui2023EfficientAE} and Tamil~\cite{Balachandran2023TamilLlamaAN} Llamas, and Swallow for Japanese~\cite{fujii2024continual}.
This line of work assumes access to a substantial amount of target language data, e.g. 312B characters of Japanese text used for Swallow.
This might not be feasible in low-resource settings with limited target language data or compute.
To the best of our knowledge, our work is the first to investigate vocabulary expansion for efficient decoder-based LLM inference in extremely low-resource settings by assuming access to a small amount of target language data (30K sentences).

\section{Problem Statement} \label{sec:problem}

The goal in this article is to expand the vocabulary of a source LLM to effectively support a target language in low-resource settings. 
This involves transitioning from a source model $\mathcal{M}_\text{s}$ (with vocabulary $\mathcal{V}_\text{s}$ and tokenizer $\mathcal{T}_\text{s}$) to a model $\mathcal{M}_\text{t}$ that supports an expanded target vocabulary $\mathcal{V}_\text{t}$ and tokenizer $\mathcal{T}_\text{t}$.

More specifically, given $\mathcal{M}_\text{s}$, $\mathcal{V}_\text{s}$, $\mathcal{T}_\text{s}$, and target language data $\mathcal{D}$, $\mathcal{M}_\text{t}$ is constructed as follows:
\begin{enumerate}
    \item \textbf{Crafting the Auxiliary Tokenizer}: A target language-specific auxiliary tokenizer is firstly trained on $\mathcal{D}$.
    This tokenizer is built from scratch using $\mathcal{D}$, allowing it to capture language-specific nuances. It comes with its own vocabulary, $\mathcal{V}_\text{aux}$.

    \item \textbf{Constructing the Target Vocabulary and Tokenizer}:
    \begin{enumerate}
        \item The new tokens $\mathcal{V}_\text{new}$ are identified by taking the top $k$ most frequent tokens from $\mathcal{V}_\text{aux}$ that are \textit{not} already present in $\mathcal{V}_\text{s}$ (i.e. $\mathcal{V}_\text{new} = \text{top } k\in\mathbb{N} \text{ tokens from } \mathcal{V}_\text{aux} \setminus (\mathcal{V}_\text{s} \cap \mathcal{V}_\text{aux})$).

        \item The target vocabulary $\mathcal{V}_\text{t}$ is then formed by combining the original source vocabulary $\mathcal{V}_\text{s}$ with these newly identified tokens: $\mathcal{V}_\text{t} = \mathcal{V}_\text{s} \cup \mathcal{V}_\text{new}$.

        \item The target tokenizer $\mathcal{T}_\text{t}$ is subsequently derived to operate on this expanded $\mathcal{V}_\text{t}$.
    \end{enumerate}

    \item \textbf{Initializing and Adapting the Target Model}:
    \begin{enumerate}
        \item The target model $\mathcal{M}_\text{t}$ begins as a copy of $\mathcal{M}_\text{s}$, inheriting its identical architecture and pre-trained weights.

        \item Its embedding and output layer matrices are then expanded to accommodate the larger $\mathcal{V}_\text{t}$. Specifically, the dimensionality of the embedding layer becomes $|\mathcal{V}_\text{t}| \times H_\text{t}$ and that of the output layer becomes $H_\text{t} \times |\mathcal{V}_\text{t}|$, where $H_\text{t}$ is the hidden dimensionality of $\mathcal{M}_\text{t}$.

        \item Each representation for a new token (i.e. tokens in $\mathcal{V}_\text{new}$) is initialized using a target parameter initialization method (detailed in \S\ref{sec:vocab_init}). If the weights of both embeddings and language modeling head (i.e. output layer) are not tied, they are initialized separately.\footnote{We follow the original configuration of the source model regarding weight tying for the embeddings and output layer to preserve its original behavior as closely as possible.}

        \item $\mathcal{M}_\text{t}$ undergoes continual pre-training (i.e. LAPT) on the target language data $\mathcal{D}$ using a causal language modeling (\textsc{clm}) objective.
        
    \end{enumerate}
\end{enumerate}

\section{Target Parameter Initialization} \label{sec:vocab_init}

After vocabulary expansion (step 2 in \S\ref{sec:problem}), the new embeddings should be initialized. Our aim is to investigate the effectiveness and robustness of different initialization approaches in low-resource settings. For that purpose, we evaluate: (1) random initialization; (2) initialization based on auxiliary models; and (3) heuristic-based initialization.\footnote{Our primary focus in the subsequent discussion is on embedding initialization for simplicity.  However, if the weights of the embeddings and the language modeling head are not tied, the language modeling head should also be initialized independently, using the identical process applied to the embeddings.}

\subsection{Random Initialization}
For new tokens, we randomly initialize the weights of their embeddings by sampling from $\mathcal{N}(\mu, \sigma^2)$.
Here, $\mu$ and $\sigma$ are the mean and standard deviation of the token embeddings from $\mathcal{M}_\text{s}$ (\textbf{Random}).
This is the most simple and common approach when adapting English-centric LLMs to a target language via vocabulary expansion in high-resource settings~\cite{Cui2023EfficientAE,Balachandran2023TamilLlamaAN,Larcher2023CabritaCT,choi-etal-2024-optimizing}.

\subsection{Initialization Based on Auxiliary Models}
\textbf{FOCUS}~\citep{dobler-de-melo-2023-focus} is a state-of-the-art CVA method that relies on auxiliary embeddings, i.e. fastText~\cite{bojanowski-etal-2017-enriching}, for initialization.\footnote{Due to resource constraints, we only report results using FOCUS as a representative initialization method that relies on auxiliary embeddings since it outperforms other similar methods such as WECHSEL~\cite{minixhofer-etal-2022-wechsel} for vocabulary replacement.} 
The main assumption is that semantic transfer of embeddings should result in better initialization over Random.
We apply FOCUS by tokenizing $\mathcal{D}$ using $\mathcal{T}_\text{t}$ and train a fastText model for each language.\footnote{Preliminary experiments with FOCUS using off-the-shelf pre-trained word-level fastText models yielded lower performance.}

\subsection{Heuristic-based Initialization}

Sophisticated methods such as FOCUS require auxiliary embeddings trained in the target language, which might not be available or hard to train in low-resource settings.
Motivated by this, we evaluate a set of heuristic-based initialization methods that do not rely on any external data or model and can be applied to any language.

\paragraph{Mean}
A straightforward approach is to initialize the weights of each new token in $\mathcal{V}_\text{new}$ by averaging the weights of their corresponding source tokens, which are identified using $\mathcal{T}_\text{s}$~\cite{yao-etal-2021-adapt}.
\citet{koto-etal-2021-indobertweet} and \citet{gee-etal-2022-fast} have also followed a similar approach for vocabulary replacement in domain adaptation.
More recently, \citet{tejaswi-etal-2024-exploring} and \citet{mundra-etal-2024-empirical} have demonstrated the effectiveness of this mean initialization for vocabulary expansion under high-resource settings (i.e. at least 200M tokens and 2.5B tokens, respectively).
However, our study specifically investigates low-resource settings with only 30K sentences, equivalent to at most 5M tokens.

\paragraph{Merge}
Mean uses solely $\mathcal{T}_\text{s}$, which might produce subtokens from $\mathcal{V}_\text{s}$ that are not semantically related to the new target token.
To overcome this issue, we propose using merge rules from $\mathcal{T}_\text{t}$ to effectively initialize $\mathcal{V}_\text{new}$.
Merge rules describe how $\mathcal{T}_\text{t}$ can combine two subtokens into one.
Through these rules, each new token in $\mathcal{V}_\text{new}$ can be decomposed into several existing subtokens from $\mathcal{V}_\text{s}$.

For instance, consider a new token: `superherohype' in $\mathcal{V}_\text{new}$.
According to the merge rules in $\mathcal{T}_\text{t}$, it can first be decomposed into (`superhero', `hype').
If `superhero` is also a new token, it can be further decomposed into (`super', `hero'). 
This process continues until all constituent parts are tokens from $\mathcal{V}_\text{s}$ (in this case, 'super', 'hero', and 'hype'). 
We hypothesize that such hierarchically derived subtokens from $\mathcal{V}_\text{s}$ are more semantically related to the new token than those obtained by simple averaging, as they leverage the specific tokenization information embedded in $\mathcal{T}_\text{t}$.

The initialization process is as follows:
\begin{enumerate}
    \item We identify all merge rules in $\mathcal{T}_\text{t}$ that result in a token found in $\mathcal{V}_\text{new}$.
    \item Using these identified merge rules, we generate a hierarchical mapping for each new target token down to its constituent source subtokens in $\mathcal{V}_\text{s}$.
    \item For each new token, we compute the hierarchical mean of the embeddings of its associated source subtokens, guided by the mapping information.
\end{enumerate}

\paragraph{Align}
Initializing the weights of new tokens in $\mathcal{V}_\text{t}$ by simply averaging the weights of their constituent subtokens from $\mathcal{T}_\text{s}$ as in Mean can be suboptimal.
This naive approach does not account for how tokenization might change in a full sequence, as opposed to a single token.

Consider the word: `\textunderscore cup' and a new token: '\textunderscore cu'.\footnote{`\textunderscore ' stands for a whitespace.}
While Mean averages the weights of `\textunderscore c' and `u' from $\mathcal{T}_\text{s}$ to initialize that of '\textunderscore cu', this can be suboptimal. 
If $\mathcal{T}_\text{s}$ tokenizes `\textunderscore cup' as `\textunderscore c' + `up', but $\mathcal{T}_\text{t}$ tokenizes it as `\textunderscore cu' + `p', `\textunderscore cu' functions as a new, distinct subword unit.
Its weight initialization should reflect this new role, perhaps focusing more on relevant overlapping components like `\textunderscore c'.

To obtain a more fine-grained semantic representation of a new token, we propose token alignment initialization that leverages mapping information between tokens tokenized with $\mathcal{T}_\text{s}$ and $\mathcal{T}_\text{t}$ and the mapping frequency (i.e. more frequent mappings are more important for the final representation). This process allows us to consider different tokenization variants for the initialization of a given token.

Specifically, the weights of each new token $t \in \mathcal{V}_\text{new}$ are initialized as follows.
\begin{enumerate}
    \item \textbf{Generate Mappings from $\mathcal{D}$}:
        \begin{enumerate}
            \item For each sentence $x \in \mathcal{D}$, we first tokenize it using both $\mathcal{T}_\text{s}$ and $\mathcal{T}_\text{t}$.

            \item We then compare these two tokenized versions to identify how each new token $t$ in $\mathcal{T}_\text{t}$'s output corresponds to sequences of subtokens from $\mathcal{T}_\text{s}$'s output.
            These correspondences are stored as a list of tuples, e.g. a token $t$ from $\mathcal{V}_\text{t}$ might map to $[(t_1, t_2), (t_3, t_4)]$ from $\mathcal{V}_\text{s}$ in different contexts.
        \end{enumerate}

    \item \textbf{Construct Unique Mappings and Frequencies}:
        \begin{enumerate}
            \item We concatenate all such mapping lists from $\mathcal{D}$ for each new token $t$.
            
            \item From this combined list, we construct a unique set of constituent tuples for $t$, denoted as $\mathbf{T}_t = (\mathbf{t}_1, \mathbf{t}_2, \mathbf{t}_3, \dots)$, where each $\mathbf{t}_i$ is a sequence of subtokens from $\mathcal{V}_\text{s}$.
            
            \item We also create a corresponding frequency list $\mathbf{F}_t = (f_{\mathbf{t}_1}, f_{\mathbf{t}_2}, f_{\mathbf{t}_3}, \dots)$, indicating how often each unique tuple $\mathbf{t}_i$ appeared in the mappings for token $t$ on $\mathcal{D}$.
        \end{enumerate}

    \item \textbf{Compute Initial Representation}:
        Finally, using $\mathbf{T}_t$ and $\mathbf{F}_t$, we aggregate the mapping information to generate the representation for token $t$ by computing 
        $\sum_{\mathbf{t} \in \mathbf{T}_t}{\left[f_{\mathbf{t}}\frac{1}{|\mathbf{t}|}\sum_{t'\in\mathbf{t}}{\mathbf{e}_{t'}}\right]}$.
        Here, $\mathbf{e}_{t'}$ is the embedding of the subtoken $t' \in \mathcal{V}_\text{s}$, $f_{\mathbf{t}}$ is the frequency of mapping $\mathbf{t}$, and $|\mathbf{t}|$ is the number of subtokens in $\mathbf{t}$.
    
\end{enumerate}

\section{Training Strategy} \label{sec:training_method}
Continual pre-training on $\mathcal{D}$ (i.e. LAPT) is an integral part of vocabulary expansion to improve the alignment of the newly initialized embeddings.
To this end, we explore: (i) the training procedure, (ii) the objective function, and (iii) the input sequence length.

\paragraph{Training Procedure}

\begin{itemize}
    \item \textbf{LoRA}: We use by default low-rank adaptation (\textbf{LoRA})~\cite{Hu2021LoRALA} applied to all linear layers, following previous work on vocabulary expansion in high-resource settings~\cite{Cui2023EfficientAE,Balachandran2023TamilLlamaAN,abbasi2023persianllamabuildingpersianlarge,choi-etal-2024-optimizing}.

    \item \textbf{2-stage}: We also consider the two-stage tuning process (\textbf{2-stage}) \cite{Cui2023EfficientAE}, where only the embeddings and language modeling head are first updated, followed by tuning the LoRA modules.
    This process can help minimize the risk of overfitting to the initial embedding state, which might be suboptimal~\cite{downey-etal-2023-embedding}.

    \item \textbf{2x2 LS}: We train only the top and bottom two layers (\textbf{2x2 LS}) of the model, following \citet{remy2024transtokenization}. This calibrates only the parts closely related to the encoding and decoding of the target language~\cite{wendler-etal-2024-llamas,tang-etal-2024-language}, minimizing changes to the source model.
    
\end{itemize}
\noindent Note that none of these approaches have ever been compared to the standard LoRA or against each other. We tune the embeddings and language modeling head in each case.

\paragraph{Objective Function}

\begin{itemize}
    \item \textbf{\textsc{clm}}: We first evaluate a causal language modeling objective (\textsc{clm}), which has been used by previous work in high-resource settings~\citep[\textit{inter alia}]{Cui2023EfficientAE,Balachandran2023TamilLlamaAN,Larcher2023CabritaCT}, as a default training objective.

    \item \textbf{\textsc{mtp}}:
    We also consider a multi-token prediction (\textsc{mtp}) objective~\cite{pmlr-v235-gloeckle24a}, where the model must predict multiple consecutive tokens at each timestep rather than a single token.
    \textsc{mtp} has been proposed for pre-training and exhibited performance gains over \textsc{clm}.
    However, it has yet to be explored for continual pre-training for cross-lingual transfer.
    We experiment with one additional language modeling head (i.e. predicting two consecutive tokens at a time) and initialize the additional weights with those from the original language modeling head.\footnote{We find in our preliminary analysis that random initialization does not work well.}
    
\end{itemize}

\paragraph{Input Sequence Length}
We shorten the default input sequence length from \textbf{2,048} to \textbf{512}, thereby increasing the number of training batches.
We hypothesize that this is critical in low-resource settings, as a small number of model updates could be prone to underfitting.

\section{Experimental Setup}
\subsection{Source Models}
We use \textbf{Llama2} 7B~\cite{Touvron2023Llama2O}, an English-centric non-instruction-tuned model, with its $\mathcal{T}_\text{s}$ based on byte-fallback Byte Pair Encoding (BPE)~\cite{sennrich-etal-2016-neural} and $|\mathcal{V}_\text{s}|$ set to 32K in the experiments.
We also use \textbf{Llama3} 8B~\cite{Dubey2024TheL3} and \textbf{Gemma2} 9B~\cite{Riviere2024Gemma2I} for consistency in our analysis. These models have a far larger vocabulary size than Llama2, i.e. 128K and 256K, respectively.
Note that Llama2 and Llama3 have untied embedding and language modeling head weights, while the original configuration of Gemma2 includes weight tying.

\subsection{Target Languages and Data}

We experiment with a typologically diverse set of ten languages with various scripts. This includes German (Indo-European) and Swahili (Niger--Congo) for the Latin script, and Arabic (Afroasiatic), Burmese (Sino-Tibetan), Greek (Indo-European), Hindi (Indo-European), Japanese (Japonic), Sinhala (Indo-European), Telugu (Dravidian), and Thai (Kra--Dai) for non-Latin scripts.
We select these languages because of the availability of downstream task datasets with the same task formulation across languages, with a particular focus on generation tasks. 

To simulate a realistic low-resource adaptation scenario, we follow \citet{yong-etal-2023-bloom} in using 30K sentences per language.
With $|\mathcal{D}|$ set to 30K, this equates to up to approximately 5M tokens.
These sentences are randomly sampled from their language-specific subcorpus of CC-100~\cite{conneau-etal-2020-unsupervised}.
Note that previous work on vocabulary expansion for generative LLMs~\cite{tejaswi-etal-2024-exploring,mundra-etal-2024-empirical} uses at least 200M tokens and 2.5B tokens respectively, which is at least 40 times larger than our training budget.\footnote{For additional context, the BabyLM challenge~\cite{warstadt-etal-2023-findings,conll-2024-babylm}, where participants must pre-train a model from scratch under low-resource settings, utilizes training budgets of up to 100M words, motivated by the observation that children are exposed to fewer than 100M words by 13 years of age.
While the BabyLM challenge is a challenging task, our setting, with access to less than 5M tokens, represents an even more extreme low-resource scenario, posing a distinct challenge for effective target language adaptation.}

\subsection{Baselines} \label{subsec:baselines}
We use the following two methods as our baselines: 
\begin{enumerate}
    \item \textbf{Source}: We use the off-the-shelf source base (i.e. non-instruction-tuned) model $\mathcal{M}_\text{s}$ without any adaptation, following \citet{tejaswi-etal-2024-exploring}. This provides a crucial reference point, allowing us to quantify the inherent performance in the target language before language-specific tuning, and thereby clearly measure specific gains from vocabulary expansion and LAPT.
    
    \item \textbf{CPT-Only}: We continue pre-training (CPT) \textit{Source} on $\mathcal{D}$, retaining its original vocabulary $\mathcal{V}_\text{s}$. 
    This differs from vocabulary expansion approaches, which expand the original vocabulary $\mathcal{V}_\text{s}$ to include new terms $\mathcal{V}_\text{t} = \mathcal{V}_\text{s} \cup \mathcal{V}_\text{new}$ as mentioned in \S\ref{sec:problem}.
\end{enumerate} 
\noindent It is important to emphasize that these baselines do not offer any inference speedups.

\subsection{Evaluation}

\paragraph{Tasks}
We use both generation and classification target language tasks to evaluate each approach.
For generation tasks, we use (1) English-to-target machine translation (\textsc{mt}) using FLORES-200~\citep{nllb-22} and (2) summarization (\textsc{sum}) including German MLSUM~\cite{scialom-etal-2020-mlsum}, GreekSUM~\cite{evdaimon2023greekbart}, and XL-Sum~\cite{hasan-etal-2021-xl} for the rest. Note that \textsc{sum} performance is not directly comparable between different languages as the data does not match across languages.
For classification tasks, we employ multiple-choice reading comprehension (\textsc{mc}) using Belebele~\citep{bandarkar-etal-2024-belebele} and Global MMLU (\textsc{gmmlu})~\citep{singh2024globalmmluunderstandingaddressing} as a general knowledge and reasoning benchmark. Note that \textsc{gmmlu} does not support Burmese and Thai.

\paragraph{Number of Samples}
Following \citet{ahia-etal-2023-languages}, we use 500 random samples for generation tasks (\textsc{sum} and \textsc{mt}). \textsc{mc} and \textsc{gmmlu} use their full test sets for evaluation. Specifically, \textsc{mc} has 900 samples per language subset, while \textsc{gmmlu} has 14K samples.

\paragraph{Prompt Templates}
For \textsc{sum} prompt templates, we translate the English templates from \citet{ahia-etal-2023-languages} using a machine translation API, as in \citet{yong-etal-2023-bloom}. For \textsc{mt}, we create an English template and then translate it into each target language.
For \textsc{mc} and \textsc{gmmlu}, we follow the default template provided by Hugging Face LightEval~\cite{lighteval}.
The complete prompt templates are listed in Table \ref{tab:prompt} in the Appendix.

\subsection{Evaluation Metrics}
\paragraph{Task Performance}
We use accuracy for \textsc{mc} and \textsc{gmmlu}, chrF~\cite{popovic-2015-chrf} for \textsc{mt}, and ROUGE-L~\cite{lin-2004-rouge} for \textsc{sum}.
In the analysis (\S\ref{subsec:llama3}), we also use BLEURT~\cite{sellam-etal-2020-bleurt} as an auxiliary metric for \textsc{sum}.

We report average zero-shot performance across five different runs for the generation tasks, namely \textsc{sum} and \textsc{mt}.
For the classification tasks, we report single-run three-shot performance for \textsc{mc} and five-shot performance for \textsc{gmmlu} as these tasks are deterministically evaluated with temperature set to zero.

\paragraph{Perplexity}
We report perplexity on 100K language-specific held-out CC-100 sentences as an auxiliary metric for evaluating model performance.

\paragraph{Inference Efficiency}
We measure inference efficiency as the number of tokens generated per second \cite{hong-etal-2024-accelerating}.

\subsection{Implementation Details}

\paragraph{Hyperparameters}
We set $|\mathcal{V}_\text{aux}|$ to 50K across languages and the number of new target tokens $|\mathcal{V}_\text{new}|$ to 100 by default. We investigate the effect of varying $|\mathcal{V}_\text{new}|$ in \S\ref{subsec:vocab_size}.
For LAPT, we train each model for two epochs with a batch size of 8, a maximum learning rate of 1e-4, and a sequence length of 2,048.
We set a LoRA rank to 8 following previous work~\cite{Cui2023EfficientAE,abbasi2023persianllamabuildingpersianlarge,Lin2024MaLA500ML}.
Table \ref{tab:hyperparams_pretraining} in the Appendix details the hyperparameter configurations employed during both the training and inference phases.

To make a fair comparison, we do not conduct any parameter tuning and use the same ones across all approaches.
For \textsc{sum}, we truncate an article whenever it exceeds the maximum prompt length of 4,096 to avoid the CUDA out-of-memory error.

\paragraph{Evaluation Metrics Computation}
To compute ROUGE-L, we split sentences with an mT5~\cite{xue-etal-2021-mt5} tokenizer as preprocessing following \citet{maynez-etal-2023-benchmarking} and subsequently call \texttt{rouge\_scorer}\footnote{\url{https://github.com/csebuetnlp/xl-sum/tree/master/multilingual_rouge_scoring}} to compute the metric.
We use BLEURT-20 to compute BLEURT.\footnote{\url{https://github.com/google-research/bleurt}}

\paragraph{Libraries and Hardware}
For Llama2 and Gemma2, we train tokenizers using SentencePiece~\cite{kudo-richardson-2018-sentencepiece} and convert them into the Hugging Face Tokenizers~\cite{Moi_HuggingFace_s_Tokenizers_2023} format.
For Llama3, we train tokenizers using Hugging Face Tokenizers.
We implement our models using PyTorch~\cite{NEURIPS2019_9015}, Hugging Face Transformers~\cite{wolf-etal-2020-transformers} and PEFT~\cite{peft}.
We preprocess datasets with Hugging Face Datasets~\cite{lhoest-etal-2021-datasets}.
For evaluation, we use Hugging Face LightEval~\cite{lighteval}.
We use either four NVIDIA V100 (32GB) or a single A100 (80GB) for LAPT. Evaluation utilizes a single NVIDIA V100 (32GB) or A100 (80GB). Each analysis utilizes a consistent hardware configuration to ensure accurate measurement of inference efficiency.

\section{Results and Analysis} \label{sec:results}

\subsection{Target Parameter Initialization} \label{subsec:vocab_init}
We analyze the effect of different target parameter initialization methods (\S\ref{sec:vocab_init}) on task performance and inference efficiency using Llama2 as source.

\subsubsection{Task Performance and Perplexity} \label{subsec:performance}
Table \ref{tab:main_result} show the performance of all methods on generation tasks, while Table \ref{tab:ppl_result} shows the corresponding perplexities on the held-out language-specific dataset.

\paragraph{Performance on Generation Tasks}
Models initialized with Mean and Align generally exhibit performance comparable to or better than Source.
Specifically, they outperform (by $>$ 2 points) or match (within 2 points) Source in 16 and 15 out of 20 cases, respectively.
Mean shows a positive gain over Source in 10 cases, while Align achieves this in 12 cases.
These results aligns with their perplexity scores (Table \ref{tab:ppl_result}), where Align generally yields the lowest perplexity across languages, closely followed by Mean.
While Merge also matches or surpasses Source performance in 16 cases (with 10 positive gains), it often shows the largest perplexities among the three heuristic-based initialization methods.
We speculate that Merge might be less informative than these two methods, as it does not rely on surface information for initialization, resulting in slightly larger perplexity.

\setlength{\tabcolsep}{4pt}
\begin{table*}[!t]
\caption{Mean performance on generation tasks (\textsc{mt} and \textsc{sum}) over five runs in low-resource settings (30\text{K} sentences) using Llama2 as source.
\colorbox{green!20}{Green} shades indicate positive performance change over Source per language and task, respectively.
}
\begin{center}
\resizebox{\linewidth}{!}{%
\begin{tabular}{llllllllllllllllllllll}
\toprule
 & & \multicolumn{2}{c}{\textbf{Arabic}}  
 & \multicolumn{2}{c}{\textbf{Burmese}} 
 & \multicolumn{2}{c}{\textbf{German}} 
 & \multicolumn{2}{c}{\textbf{Greek}} 
 & \multicolumn{2}{c}{\textbf{Hindi}} 
 & \multicolumn{2}{c}{\textbf{Japanese}}
 & \multicolumn{2}{c}{\textbf{Sinhala}} 
 & \multicolumn{2}{c}{\textbf{Swahili}} 
 & \multicolumn{2}{c}{\textbf{Telugu}} 
 & \multicolumn{2}{c}{\textbf{Thai}}\\

 & & \multicolumn{2}{c}{\scriptsize Afroasiatic}  
 & \multicolumn{2}{c}{\scriptsize Sino-Tibetan}  
 & \multicolumn{2}{c}{\scriptsize Indo-European}
 & \multicolumn{2}{c}{\scriptsize Indo-European}
 & \multicolumn{2}{c}{\scriptsize Indo-European}
 & \multicolumn{2}{c}{\scriptsize Japonic}
 & \multicolumn{2}{c}{\scriptsize Indo-European}
 & \multicolumn{2}{c}{\scriptsize Niger--Congo}
 & \multicolumn{2}{c}{\scriptsize Dravidian}
 & \multicolumn{2}{c}{\scriptsize Kra--Dai}\\

\cmidrule(lr){3-4}
\cmidrule(lr){5-6}
\cmidrule(lr){7-8}
\cmidrule(lr){9-10}
\cmidrule(lr){11-12}
\cmidrule(lr){13-14}
\cmidrule(lr){15-16}
\cmidrule(lr){17-18}
\cmidrule(lr){19-20}
\cmidrule(lr){21-22}

& \textbf{Model} & \multicolumn{1}{c}{\scriptsize \textsc{mt}} & \multicolumn{1}{c}{\scriptsize \textsc{sum}} 
& \multicolumn{1}{c}{\scriptsize \textsc{mt}} & \multicolumn{1}{c}{\scriptsize \textsc{sum}} 
& \multicolumn{1}{c}{\scriptsize \textsc{mt}} & \multicolumn{1}{c}{\scriptsize \textsc{sum}} 
& \multicolumn{1}{c}{\scriptsize \textsc{mt}} & \multicolumn{1}{c}{\scriptsize \textsc{sum}} 
& \multicolumn{1}{c}{\scriptsize \textsc{mt}} & \multicolumn{1}{c}{\scriptsize \textsc{sum}} 
& \multicolumn{1}{c}{\scriptsize \textsc{mt}} & \multicolumn{1}{c}{\scriptsize \textsc{sum}} 
& \multicolumn{1}{c}{\scriptsize \textsc{mt}} & \multicolumn{1}{c}{\scriptsize \textsc{sum}} 
& \multicolumn{1}{c}{\scriptsize \textsc{mt}} & \multicolumn{1}{c}{\scriptsize \textsc{sum}} 
& \multicolumn{1}{c}{\scriptsize \textsc{mt}} & \multicolumn{1}{c}{\scriptsize \textsc{sum}} 
& \multicolumn{1}{c}{\scriptsize \textsc{mt}} & \multicolumn{1}{c}{\scriptsize \textsc{sum}} \\

\midrule
\rowcolor{gray!25}
& Source & .08\textsubscript{{\tiny.03}} & 36\textsubscript{{\tiny0.0}} & .03\textsubscript{{\tiny.00}} & 24\textsubscript{{\tiny0.1}} & .25\textsubscript{{\tiny.01}} & 25\textsubscript{{\tiny0.1}} & .08\textsubscript{{\tiny.03}} & 25\textsubscript{{\tiny0.2}} & .12\textsubscript{{\tiny.03}} & 39\textsubscript{{\tiny0.2}} & .03\textsubscript{{\tiny.03}} & 22\textsubscript{{\tiny0.1}} & .05\textsubscript{{\tiny.00}} & 26\textsubscript{{\tiny0.1}} & .01\textsubscript{{\tiny.00}} & 27\textsubscript{{\tiny0.4}} & .06\textsubscript{{\tiny.00}} & 21\textsubscript{{\tiny0.2}} & .06\textsubscript{{\tiny.02}} & 22\textsubscript{{\tiny0.2}}\\

& CPT-only & \cellcolor{green!20}.18\textsubscript{{\tiny.00}} & 36\textsubscript{{\tiny0.2}} & \cellcolor{green!20}.11\textsubscript{{\tiny.00}} & 24\textsubscript{{\tiny0.3}} & .20\textsubscript{{\tiny.00}} & 21\textsubscript{{\tiny0.1}} & \cellcolor{green!20}.21\textsubscript{{\tiny.00}} & 25\textsubscript{{\tiny0.3}} & \cellcolor{green!20}.21\textsubscript{{\tiny.00}} & 38\textsubscript{{\tiny0.2}} & \cellcolor{green!20}.08\textsubscript{{\tiny.00}} & \cellcolor{green!20}23\textsubscript{{\tiny0.2}} & \cellcolor{green!20}.07\textsubscript{{\tiny.00}} & 23\textsubscript{{\tiny0.1}} & \cellcolor{green!20}.12\textsubscript{{\tiny.00}} & \cellcolor{green!20}30\textsubscript{{\tiny0.1}} & \cellcolor{green!20}.09\textsubscript{{\tiny.00}} & \cellcolor{green!20}29\textsubscript{{\tiny0.1}} & \cellcolor{green!20}.12\textsubscript{{\tiny.00}} & \cellcolor{green!20}23\textsubscript{{\tiny0.2}}\\

\midrule
\multirow{5}{*}{\rotatebox[origin=c]{90}{\parbox[c]{2cm}{\centering \textbf{+Speedup}}}} 

& Random & .07\textsubscript{{\tiny.00}} & 32\textsubscript{{\tiny0.3}} & \cellcolor{green!20}.05\textsubscript{{\tiny.00}} & 17\textsubscript{{\tiny0.3}} & \cellcolor{green!20}.28\textsubscript{{\tiny.00}} & 23\textsubscript{{\tiny0.3}} & \cellcolor{green!20}.13\textsubscript{{\tiny.00}} & 20\textsubscript{{\tiny0.1}} & .09\textsubscript{{\tiny.00}} & 35\textsubscript{{\tiny0.2}} & \cellcolor{green!20}.18\textsubscript{{\tiny.00}} & 22\textsubscript{{\tiny0.2}} & \cellcolor{green!20}.07\textsubscript{{\tiny.00}} & \cellcolor{green!20}29\textsubscript{{\tiny0.1}} & \cellcolor{green!20}.13\textsubscript{{\tiny.00}} & \cellcolor{green!20}\cellcolor{green!20}29\textsubscript{{\tiny0.1}} & .06\textsubscript{{\tiny.00}} & \cellcolor{green!20}26\textsubscript{{\tiny0.2}} & \cellcolor{green!20}.07\textsubscript{{\tiny.00}} & 19\textsubscript{{\tiny0.2}}\\

& FOCUS & .07\textsubscript{{\tiny.00}} & 32\textsubscript{{\tiny0.2}} & .03\textsubscript{{\tiny.00}} & 16\textsubscript{{\tiny0.3}} & \cellcolor{green!20}.28\textsubscript{{\tiny.00}} & 22\textsubscript{{\tiny0.4}} & \cellcolor{green!20}.17\textsubscript{{\tiny.00}} & 18\textsubscript{{\tiny0.1}} & .10\textsubscript{{\tiny.00}} & 34\textsubscript{{\tiny0.2}} & \cellcolor{green!20}.19\textsubscript{{\tiny.00}} & 22\textsubscript{{\tiny0.2}} & .05\textsubscript{{\tiny.00}} & \cellcolor{green!20}29\textsubscript{{\tiny0.3}} & \cellcolor{green!20}.18\textsubscript{{\tiny.00}} & \cellcolor{green!20}29\textsubscript{{\tiny0.1}} & .05\textsubscript{{\tiny.00}} & \cellcolor{green!20}26\textsubscript{{\tiny0.2}} & .06\textsubscript{{\tiny.00}} & 17\textsubscript{{\tiny0.1}}\\

\cmidrule{2-22}

& Mean & .06\textsubscript{{\tiny.00}} & 33\textsubscript{{\tiny0.1}} & \cellcolor{green!20}.04\textsubscript{{\tiny.00}} & 20\textsubscript{{\tiny0.4}} & .24\textsubscript{{\tiny.01}} & 23\textsubscript{{\tiny0.2}} & \cellcolor{green!20}.14\textsubscript{{\tiny.00}} & 19\textsubscript{{\tiny0.1}} & .12\textsubscript{{\tiny.00}} & 34\textsubscript{{\tiny0.2}} & \cellcolor{green!20}.20\textsubscript{{\tiny.00}} & \cellcolor{green!20}23\textsubscript{{\tiny0.1}} & \cellcolor{green!20}.06\textsubscript{{\tiny.00}} & \cellcolor{green!20}30\textsubscript{{\tiny0.3}} & \cellcolor{green!20}.12\textsubscript{{\tiny.00}} & \cellcolor{green!20}29\textsubscript{{\tiny0.1}} & .06\textsubscript{{\tiny.00}} & \cellcolor{green!20}27\textsubscript{{\tiny0.1}} & \cellcolor{green!20}.09\textsubscript{{\tiny.00}} & 22\textsubscript{{\tiny0.1}}\\

& Merge & .07\textsubscript{{\tiny.00}} & 33\textsubscript{{\tiny0.1}} & \cellcolor{green!20}.04\textsubscript{{\tiny.00}} & 9\textsubscript{{\tiny0.5}} & .25\textsubscript{{\tiny.00}} & 23\textsubscript{{\tiny0.2}} & \cellcolor{green!20}.15\textsubscript{{\tiny.00}} & 18\textsubscript{{\tiny0.3}} & .10\textsubscript{{\tiny.00}} & 34\textsubscript{{\tiny0.2}} & \cellcolor{green!20}.18\textsubscript{{\tiny.00}} & \cellcolor{green!20}23\textsubscript{{\tiny0.3}} & \cellcolor{green!20}.09\textsubscript{{\tiny.00}} & \cellcolor{green!20}30\textsubscript{{\tiny0.2}} & \cellcolor{green!20}.13\textsubscript{{\tiny.00}} & \cellcolor{green!20}29\textsubscript{{\tiny0.1}} & .05\textsubscript{{\tiny.00}} & \cellcolor{green!20}26\textsubscript{{\tiny0.3}} & \cellcolor{green!20}.07\textsubscript{{\tiny.00}} & 21\textsubscript{{\tiny0.2}}\\

& Align & .06\textsubscript{{\tiny.00}} & 33\textsubscript{{\tiny0.1}} & \cellcolor{green!20}.04\textsubscript{{\tiny.00}} & 15\textsubscript{{\tiny0.7}} & \cellcolor{green!20}.26\textsubscript{{\tiny.01}} & 21\textsubscript{{\tiny0.2}} & \cellcolor{green!20}.16\textsubscript{{\tiny.00}} & 17\textsubscript{{\tiny0.2}} & \cellcolor{green!20}.13\textsubscript{{\tiny.00}} & 35\textsubscript{{\tiny0.2}} & \cellcolor{green!20}.20\textsubscript{{\tiny.00}} & \cellcolor{green!20}23\textsubscript{{\tiny0.1}} & \cellcolor{green!20}.08\textsubscript{{\tiny.00}} & \cellcolor{green!20}30\textsubscript{{\tiny0.3}} & \cellcolor{green!20}.17\textsubscript{{\tiny.00}} & \cellcolor{green!20}29\textsubscript{{\tiny0.0}} & .06\textsubscript{{\tiny.00}} & \cellcolor{green!20}27\textsubscript{{\tiny0.2}} & \cellcolor{green!20}.07\textsubscript{{\tiny.00}} & 22\textsubscript{{\tiny0.2}}\\

\bottomrule
\end{tabular}%
}
\label{tab:main_result}
\end{center}
\end{table*}

\setlength{\tabcolsep}{3pt}
\renewcommand*{\arraystretch}{1.0}
\begin{table}[!t]

\caption{Perplexity on language-specific held-out dataset using Llama2 as source. Note that results are not comparable between models with \colorbox{gray!25}{gray} and others due to their difference in vocabulary. \textbf{Bold} and \underline{underlined} indicate the best and second-best perplexities among adapted models for each language.}

\begin{center}
\small
\begin{tabular}{lllllllllll}
\toprule
\textbf{Model} & \textbf{ar} & \textbf{my} & \textbf{de} & \textbf{el} & \textbf{hi} & \textbf{ja} & \textbf{si} & \textbf{sw} & \textbf{te} & \textbf{th}\\

\midrule
\rowcolor{gray!25}
Source  & 8.3 & 5.0 & 35.7 & 4.8 & 6.4 & 20.4 & 3.5 & 47.2 & 2.4 & 9.4\\
\rowcolor{gray!25}
CPT-only & 4.2 & 2.7 & 10.9 & 2.9 & 3.2 & 5.4 & 2.3 & 12.7 & 1.8 & 4.3\\
\midrule
Random & 11.9 & 11.9 & 12.0 & 7.1 & 8.3 & \underline{15.0} & 8.7 & 13.9 & 7.1 & 8.8\\
FOCUS & 11.5 & 13.8 & 12.1 & 6.6 & 8.9 & 14.8 & 9.3 & \underline{13.7} & 7.9 & 9.4\\
\midrule
Mean & \underline{9.4} & \underline{11.6} & \textbf{11.7} & \underline{6.0} & \underline{6.4} & \textbf{14.7} & \underline{8.1} & \textbf{13.5} & \underline{6.9} & \underline{7.8}\\
Merge & 9.8 & 12.6 & \underline{11.8} & 6.1 & 6.7 & 15.1 & 8.7 & \underline{13.7} & 7.3 & 8.0\\
Align & \textbf{9.3} & \textbf{11.2} & \textbf{11.7} & \textbf{5.9} & \textbf{6.3} & 15.1 & \textbf{8.0} & \textbf{13.5} & \textbf{6.7} & \textbf{7.6}\\
\bottomrule
\end{tabular}%
\label{tab:ppl_result}
\end{center}
\end{table}

The baseline Random and FOCUS models perform similarly (within 2 points) to or better than Source in 14 cases, with 10 and 7 cases showing positive gains, respectively.
These results place them among the least effective approaches.
FOCUS, in particular, shows the fewest cases with positive gains and the worst perplexity in six languages (i.e. Burmese, German, Hindi, Sinhala, Telugu, and Thai).
These findings suggest that sophisticated initialization methods do not always guarantee superior performance, possibly due to underfitting of an auxiliary embedding model. Further, the popular Random approach is not always optimal in low-resource scenarios.
Thus, we conclude that \textit{models initialized with Mean and Align are more likely to perform competitively with Source on generation tasks.}

Analyzing performance by language and task, Mean and Align models show positive gains over Source in six and eight out of ten languages for \textsc{mt}, respectively.
In cases where gains are not observed, any performance degradation is negligible (within 2 points). 
Notably, they substantially outperform Source in Greek, Japanese, and Swahili \textsc{mt}, with improvements ranging from 8 to 17 points.

However, for \textsc{sum}, these adapted models often do not yield a performance gain over Source in the majority of the languages, except for Japanese, Sinhala, Swahili, and Telugu.
In particular, Burmese and Greek show a substantial performance drop of up to 9 and 8 points, respectively.
We hypothesize that \textsc{sum} may necessitate more training data than \textsc{mt} because generating longer text (up to 128 tokens, using long context) requires models to have strong generative capabilities in the target language. 
We later address this challenge in \S\ref{subsec:training_method} by showing that these performance gaps can be drastically narrowed using our alternative training strategies, further contributing to the competitiveness of vocabulary expansion approaches to Source.

Turning to CPT-only (i.e. continual pre-training without vocabulary expansion), it demonstrates strong performance, matching or improving upon Source in 17 out of 20 cases, a higher count by at least one case than achieved by the models with the heuristic-based initialization.
A direct comparison against the best-performing Align\footnote{It is best-performing because it generally achieves the lowest perplexities across languages and is the most likely among different initialization methods to provide a positive gain over Source.} further confirms this.
CPT-only outperforms Align in 10 out of 20 cases, whereas Align only wins in four. 
These results demonstrate the overall superiority of CPT-only over vocabulary expansion approaches.
This trend aligns with previous work on CVA~\cite{downey-etal-2023-embedding,yamaguchi-etal-2024-empirical} suggesting that CPT-only can often perform better than vocabulary expansion approaches in low-resource settings, possibly due to its reliance on robust and well-aligned original embeddings.

\setlength{\tabcolsep}{4pt}
\begin{table*}[!t]
\caption{Mean performance on classification tasks (\textsc{mc} and \textsc{gmmlu}) in low-resource settings (30\text{K} sentences) using Llama2 as source.
\colorbox{green!20}{Green} shades indicate positive performance change over Source per language and task, respectively.
Note that \textsc{gmmlu} does not support Burmese and Thai.
}
\begin{center}
\resizebox{\linewidth}{!}{%
\begin{tabular}{llllllllllllllllllllll}
\toprule
 & & \multicolumn{2}{c}{\textbf{Arabic}}  
 & \multicolumn{2}{c}{\textbf{Burmese}} 
 & \multicolumn{2}{c}{\textbf{German}} 
 & \multicolumn{2}{c}{\textbf{Greek}} 
 & \multicolumn{2}{c}{\textbf{Hindi}} 
 & \multicolumn{2}{c}{\textbf{Japanese}}
 & \multicolumn{2}{c}{\textbf{Sinhala}} 
 & \multicolumn{2}{c}{\textbf{Swahili}} 
 & \multicolumn{2}{c}{\textbf{Telugu}} 
 & \multicolumn{2}{c}{\textbf{Thai}}\\

\cmidrule(lr){3-4}
\cmidrule(lr){5-6}
\cmidrule(lr){7-8}
\cmidrule(lr){9-10}
\cmidrule(lr){11-12}
\cmidrule(lr){13-14}
\cmidrule(lr){15-16}
\cmidrule(lr){17-18}
\cmidrule(lr){19-20}
\cmidrule(lr){21-22}
 
&  & \multicolumn{1}{c}{\scriptsize \textsc{mc}} & \multicolumn{1}{c}{\scriptsize \textsc{gmmlu}} 
& \multicolumn{1}{c}{\scriptsize \textsc{mc}} & \multicolumn{1}{c}{\scriptsize \textsc{gmmlu}} 
& \multicolumn{1}{c}{\scriptsize \textsc{mc}} & \multicolumn{1}{c}{\scriptsize \textsc{gmmlu}} 
& \multicolumn{1}{c}{\scriptsize \textsc{mc}} & \multicolumn{1}{c}{\scriptsize \textsc{gmmlu}} 
& \multicolumn{1}{c}{\scriptsize \textsc{mc}} & \multicolumn{1}{c}{\scriptsize \textsc{gmmlu}} 
& \multicolumn{1}{c}{\scriptsize \textsc{mc}} & \multicolumn{1}{c}{\scriptsize \textsc{gmmlu}} 
& \multicolumn{1}{c}{\scriptsize \textsc{mc}} & \multicolumn{1}{c}{\scriptsize \textsc{gmmlu}} 
& \multicolumn{1}{c}{\scriptsize \textsc{mc}} & \multicolumn{1}{c}{\scriptsize \textsc{gmmlu}} 
& \multicolumn{1}{c}{\scriptsize \textsc{mc}} & \multicolumn{1}{c}{\scriptsize \textsc{gmmlu}} 
& \multicolumn{1}{c}{\scriptsize \textsc{mc}} & \multicolumn{1}{c}{\scriptsize \textsc{gmmlu}} \\
\midrule

\rowcolor{gray!25}
& Source & .29 & .29 & .26 & -  & .43 & .39 & .27 & .28 & .25 & .28 & .40 & .33 & .24 & .27 & .31 & .28 & .28 & .27 & .29 & - \\

& CPT-only & \cellcolor{green!20}.30 & .29 & .22 & -  & .42 & .38 & \cellcolor{green!20}.29 & \cellcolor{green!20}.29 & \cellcolor{green!20}.28 & .28 & .39 & \cellcolor{green!20}.34 & .23 & .27 & \cellcolor{green!20}.33 & .27 & .24 & .27 & .28 & - \\
\midrule
\multirow{5}{*}{\rotatebox[origin=c]{90}{\parbox[c]{2cm}{\centering \textbf{+Speedup}}}}
& Random & .28 & .25 & .22 & -  & .42 & .38 & .22 & .27 & \cellcolor{green!20}.27 & .26 & .35 & .30 & .22 & .27 & .29 & .25 & .28 & .26 & .28 & - \\
& FOCUS & .29 & .26 & \cellcolor{green!20}.28 & -  & .41 & .38 & .24 & .26 & \cellcolor{green!20}.28 & .26 & .36 & .31 & \cellcolor{green!20}.29 & .26 & .28 & .25 & .24 & .27 & .28 & - \\
\cmidrule{2-22}
& Mean & .28 & .25 & .23 & -  & .42 & .38 & \cellcolor{green!20}.30 & .26 & \cellcolor{green!20}.27 & .26 & .38 & .29 & \cellcolor{green!20}.28 & .26 & .29 & .25 & .23 & .27 & \cellcolor{green!20}.30 & - \\
& Merge & .29 & .25 & .25 & -  & .43 & .38 & .27 & .28 & \cellcolor{green!20}.26 & .28 & .37 & .29 & \cellcolor{green!20}.27 & .26 & .28 & .25 & .28 & .26 & .29 & - \\
& Align & .28 & .25 & .26 & -  & .40 & .37 & \cellcolor{green!20}.31 & .26 & \cellcolor{green!20}.27 & .27 & .37 & .28 & \cellcolor{green!20}.27 & .27 & .29 & .25 & .22 & .27 & \cellcolor{green!20}.31 & - \\
\bottomrule

\end{tabular}%
}
\label{tab:main_result_cls}
\end{center}
\end{table*}

\paragraph{Performance on Classification Tasks}
Table \ref{tab:main_result_cls} presents the performance of all methods on classification tasks.
A distinct trend emerges compared to generation tasks: \textit{adapted models rarely yield a positive performance gain over Source.} While all vocabulary expansion approaches exhibit competitive or better performance than Source in at least 12 cases, their positive gains are limited to a maximum of four cases (22\%) (Mean and Align). This is far fewer than the maximum of 12 cases (60\%) observed on the generation tasks.

Notably, CPT-only performs on par with or outperforms Source in 16 out of 18 cases. Nonetheless, this contrasts with its strong performance on generation tasks: it shows positive gains in only 6 out of 18 cases (33\%), compared to 65\% on generation tasks.
These results suggest that while continual pre-training (i.e. LAPT) can offer benefits, its impact on discriminative tasks in low-resource settings is limited and differs from its effect on generation tasks.

We hypothesize these differences arise because our adaptation process (LAPT) primarily optimizes a causal language modeling objective on unlabeled target language data $\mathcal{D}$. While this objective is highly beneficial for generative tasks requiring fluency and extended text production, it may not directly translate to immediate gains on classification tasks.
Classification tasks, in contrast, often heavily depend on the inherent semantic and factual knowledge of a model, frequently requiring only a single token generation for prediction. Therefore, while previous work on vocabulary expansion with LAPT \cite{tejaswi-etal-2024-exploring,Cui2023EfficientAE,Balachandran2023TamilLlamaAN,choi-etal-2024-optimizing,yamaguchi2025elchatadaptingchatlanguage} often reports performance improvements in both generation and classification tasks in resource-rich settings, our findings suggest that under extremely low-resource settings (i.e. 30K sentences per language, approximately up to 5M tokens), LAPT does not inherently guarantee an improvement in these discriminative capabilities.

\setlength{\tabcolsep}{2.5pt}
\begin{table*}[t]
    \caption{Inference efficiency for: (a) generation tasks (\textsc{mt} and \textsc{sum}) over five runs; and (b) classification tasks (\textsc{mc} and \textsc{gmmlu}). We measure inference efficiency by the number of tokens generated per second (T/S). 
    The Source row shows both raw T/S and baseline 1.00x speedup per language and task.
    \textbf{Bold} and \underline{underlined} indicate the best and second-best speedups across tasks and models with the same base model for each language.}
        
    \small
    \resizebox{\linewidth}{!}{%
    \begin{tabular}{lllllllllllllllllllll}
        \multicolumn{21}{c}{(a) Generation tasks}\\
        \toprule
         \multirow{2}{*}{\textbf{Model}}
         & \multicolumn{2}{c}{\textbf{Arabic}}  
         & \multicolumn{2}{c}{\textbf{Burmese}} 
         & \multicolumn{2}{c}{\textbf{German}} 
         & \multicolumn{2}{c}{\textbf{Greek}} 
         & \multicolumn{2}{c}{\textbf{Hindi}}
         & \multicolumn{2}{c}{\textbf{Japanese}}
         & \multicolumn{2}{c}{\textbf{Sinhala}} 
         & \multicolumn{2}{c}{\textbf{Swahili}} 
         & \multicolumn{2}{c}{\textbf{Telugu}} 
         & \multicolumn{2}{c}{\textbf{Thai}}\\

        \cmidrule(lr){2-3}
        \cmidrule(lr){4-5}
        \cmidrule(lr){6-7}
        \cmidrule(lr){8-9}
        \cmidrule(lr){10-11}
        \cmidrule(lr){12-13}
        \cmidrule(lr){14-15}
        \cmidrule(lr){16-17}
        \cmidrule(lr){18-19}
        \cmidrule(lr){20-21}
        
         & \multicolumn{1}{c}{\scriptsize \textsc{mt}} & \multicolumn{1}{c}{\scriptsize \textsc{sum}}
         & \multicolumn{1}{c}{\scriptsize \textsc{mt}} & \multicolumn{1}{c}{\scriptsize \textsc{sum}}
         & \multicolumn{1}{c}{\scriptsize \textsc{mt}} & \multicolumn{1}{c}{\scriptsize \textsc{sum}}
         & \multicolumn{1}{c}{\scriptsize \textsc{mt}} & \multicolumn{1}{c}{\scriptsize \textsc{sum}}
         & \multicolumn{1}{c}{\scriptsize \textsc{mt}} & \multicolumn{1}{c}{\scriptsize \textsc{sum}}
         & \multicolumn{1}{c}{\scriptsize \textsc{mt}} & \multicolumn{1}{c}{\scriptsize \textsc{sum}}
         & \multicolumn{1}{c}{\scriptsize \textsc{mt}} & \multicolumn{1}{c}{\scriptsize \textsc{sum}}
         & \multicolumn{1}{c}{\scriptsize \textsc{mt}} & \multicolumn{1}{c}{\scriptsize \textsc{sum}}
         & \multicolumn{1}{c}{\scriptsize \textsc{mt}} & \multicolumn{1}{c}{\scriptsize \textsc{sum}}
         & \multicolumn{1}{c}{\scriptsize \textsc{mt}} & \multicolumn{1}{c}{\scriptsize \textsc{sum}}\\
        
        \midrule
        \rowcolor{gray!25}
         &  42.2 &  20.1 &  28.6 &  15.0 &  42.6 &  24.8 &  42.4 &  13.3 &  42.2 &  21.5 &  42.8 &  21.9 &  28.1 &  14.8 &  43.0 &  24.1 &  27.7 &  10.0 &  42.6 &  17.9\\
        \rowcolor{gray!25}
        \multirow{-2}{*}{Source} &  1.00{\small x} &  1.00{\small x} &   1.00{\small x} &  1.00{\small x} &  \textbf{1.00}{\small x} &  1.00{\small x} &  1.00{\small x} &  1.00{\small x} &  1.00{\small x} &  1.00{\small x} &  1.00{\small x} &  1.00{\small x} &  1.00{\small x} &  1.00{\small x} &  \textbf{1.00}{\small x} &  1.00{\small x} &  1.00{\small x} &  1.00{\small x} &  1.00{\small x} &  1.00{\small x}\\

        \midrule
        CPT-only &  1.01{\small x} &  1.13{\small x} &  0.98{\small x} &  1.01{\small x} &  \underline{0.97{\small x}} &  \textbf{1.05}{\small x} &  1.00{\small x} &  0.97{\small x} &  1.00{\small x} &  1.01{\small x} &  0.98{\small x} &  1.03{\small x} &  1.00{\small x} &  1.00{\small x} &  \underline{0.98{\small x}} &  1.21{\small x} &  0.98{\small x} &  0.99{\small x} &  0.98{\small x} &  1.01{\small x}\\
        \midrule
        Random &  \textbf{1.17}{\small x} &  1.88{\small x} &  \textbf{1.18}{\small x} &  1.48{\small x} &  0.95{\small x} &  0.97{\small x} &  1.32{\small x} &  \textbf{1.79}{\small x} &  \textbf{1.19}{\small x} &  1.57{\small x} &  1.04{\small x} &  \textbf{1.13}{\small x} &  \underline{1.56{\small x}} &  2.54{\small x} &  0.92{\small x} &  \textbf{1.27}{\small x} &  \textbf{1.93}{\small x} &  4.26{\small x} &  1.06{\small x} &  1.55{\small x}\\
        
        FOCUS &  1.14{\small x} &  1.79{\small x} &  \underline{1.13{\small x}} &  1.38{\small x} &  0.89{\small x} &  0.94{\small x} &  \textbf{1.41}{\small x} &  1.66{\small x} &  1.11{\small x} &  1.49{\small x} &  1.04{\small x} &  1.11{\small x} &  1.13{\small x} &  2.40{\small x} &  0.93{\small x} &  \underline{1.26{\small x}} &  1.79{\small x} &  4.26{\small x} &  1.06{\small x} &  1.44{\small x}\\
        
        Mean &  1.08{\small x} &  \textbf{1.99}{\small x} &  1.02{\small x} &  \textbf{2.15}{\small x} &  0.95{\small x} &  0.95{\small x} &  1.32{\small x} &  \underline{1.73{\small x}} &  \underline{1.15{\small x}} &  \underline{1.70{\small x}} &  \underline{1.08{\small x}} &  \textbf{1.13}{\small x} &  1.29{\small x} &  \textbf{2.67}{\small x} &  0.91{\small x} &  1.25{\small x} &  \underline{1.84{\small x}} &  \textbf{4.58}{\small x} &  \textbf{1.10}{\small x} &  \underline{1.62{\small x}}\\
        
        Merge &  \textbf{1.17}{\small x} &  1.95{\small x} &  1.08{\small x} &  1.20{\small x} &  0.94{\small x} &  1.00{\small x} &  \underline{1.40{\small x}} &  1.70{\small x} &  1.07{\small x} &  1.57{\small x} &  1.05{\small x} &  1.12{\small x} &  \textbf{1.58}{\small x} &  2.49{\small x} &  0.92{\small x} &  1.25{\small x} &  1.71{\small x} &  4.38{\small x} &  1.05{\small x} &  1.59{\small x}\\
        
        Align &  1.10{\small x} &  \underline{1.98{\small x}} &  1.08{\small x} &  \underline{1.90{\small x}} &  0.95{\small x} &  \underline{1.04{\small x}} &  1.37{\small x} &  1.67{\small x} &  \underline{1.15{\small x}} &  \textbf{1.77}{\small x} &  \textbf{1.09}{\small x} &  \textbf{1.13}{\small x} &  1.47{\small x} &  \underline{2.58{\small x}} &  0.93{\small x} &  \underline{1.26{\small x}} &  \underline{1.84{\small x}} &  \underline{4.42{\small x}} &  \underline{1.08{\small x}} &  \textbf{1.63}{\small x}\\
        \bottomrule
        \\
        \multicolumn{21}{c}{(b) Classification tasks}\\
        \toprule
         \multirow{2}{*}{\textbf{Model}}
         & \multicolumn{2}{c}{\textbf{Arabic}}  
         & \multicolumn{2}{c}{\textbf{Burmese}} 
         & \multicolumn{2}{c}{\textbf{German}} 
         & \multicolumn{2}{c}{\textbf{Greek}} 
         & \multicolumn{2}{c}{\textbf{Hindi}}
         & \multicolumn{2}{c}{\textbf{Japanese}}
         & \multicolumn{2}{c}{\textbf{Sinhala}} 
         & \multicolumn{2}{c}{\textbf{Swahili}} 
         & \multicolumn{2}{c}{\textbf{Telugu}} 
         & \multicolumn{2}{c}{\textbf{Thai}}\\

         \cmidrule(lr){2-3}
        \cmidrule(lr){4-5}
        \cmidrule(lr){6-7}
        \cmidrule(lr){8-9}
        \cmidrule(lr){10-11}
        \cmidrule(lr){12-13}
        \cmidrule(lr){14-15}
        \cmidrule(lr){16-17}
        \cmidrule(lr){18-19}
        \cmidrule(lr){20-21}
        
         & \multicolumn{1}{c}{\scriptsize \textsc{mc}} & \multicolumn{1}{c}{\scriptsize \textsc{gmmlu}} 
& \multicolumn{1}{c}{\scriptsize \textsc{mc}} & \multicolumn{1}{c}{\scriptsize \textsc{gmmlu}} 
& \multicolumn{1}{c}{\scriptsize \textsc{mc}} & \multicolumn{1}{c}{\scriptsize \textsc{gmmlu}} 
& \multicolumn{1}{c}{\scriptsize \textsc{mc}} & \multicolumn{1}{c}{\scriptsize \textsc{gmmlu}} 
& \multicolumn{1}{c}{\scriptsize \textsc{mc}} & \multicolumn{1}{c}{\scriptsize \textsc{gmmlu}} 
& \multicolumn{1}{c}{\scriptsize \textsc{mc}} & \multicolumn{1}{c}{\scriptsize \textsc{gmmlu}} 
& \multicolumn{1}{c}{\scriptsize \textsc{mc}} & \multicolumn{1}{c}{\scriptsize \textsc{gmmlu}} 
& \multicolumn{1}{c}{\scriptsize \textsc{mc}} & \multicolumn{1}{c}{\scriptsize \textsc{gmmlu}} 
& \multicolumn{1}{c}{\scriptsize \textsc{mc}} & \multicolumn{1}{c}{\scriptsize \textsc{gmmlu}} 
& \multicolumn{1}{c}{\scriptsize \textsc{mc}} & \multicolumn{1}{c}{\scriptsize \textsc{gmmlu}} \\
    
        \midrule
        \rowcolor{gray!25}
         & 19.4 & 29.9 & 11.1 & - & 36.6 & 34.3 & 14.0 & 23.4 & 16.9 & 24.1 & 33.4 & 34.0 & 10.9 & 15.2 & 35.4 & 33.1 & 11.1 & 12.8 & 16.7 & - \\
        \rowcolor{gray!25}
        \multirow{-2}{*}{Source} &  1.00{\small x} &  1.00{\small x} &   1.00{\small x} & - &  1.00{\small x} &  1.00{\small x} &  1.00{\small x} &  1.00{\small x} &  1.00{\small x} &  1.00{\small x} &  1.00{\small x} &  \textbf{1.00}{\small x} &  1.00{\small x} &  1.00{\small x} &  1.00{\small x} &  1.00{\small x} &  1.00{\small x} &  1.00{\small x} &  1.00{\small x} &  -\\

        \midrule
CPT-only & 1.17{\small x} & 0.98{\small x} & 1.24{\small x} & - & \textbf{1.04}{\small x} & 0.97{\small x} & 1.22{\small x} & 0.92{\small x} & 1.17{\small x} & 0.99{\small x} & \textbf{1.03}{\small x} & 0.93{\small x} & 1.27{\small x} & 1.10{\small x} & 1.04{\small x} & 1.02{\small x} & 1.02{\small x} & 1.11{\small x} & 1.20{\small x} & - \\
\midrule
Random & \textbf{1.71}{\small x} & \textbf{1.12}{\small x} & 2.12{\small x} & - & \underline{1.03}{\small x} & 1.01{\small x} & \textbf{1.68}{\small x} & \textbf{1.29}{\small x} & \textbf{1.71}{\small x} & \textbf{1.25}{\small x} & 1.02{\small x} & 0.99{\small x} & \underline{2.10}{\small x} & \underline{1.92}{\small x} & 1.05{\small x} & 1.01{\small x} & \textbf{1.84}{\small x} & 2.10{\small x} & \textbf{1.42}{\small x} & - \\
FOCUS & \textbf{1.71}{\small x} & \textbf{1.12}{\small x} & \textbf{2.13}{\small x} & - & 0.99{\small x} & 1.01{\small x} & \textbf{1.68}{\small x} & \textbf{1.29}{\small x} & \textbf{1.71}{\small x} & 1.23{\small x} & 1.02{\small x} & 0.99{\small x} & \textbf{2.11}{\small x} & \underline{1.92}{\small x} & 1.05{\small x} & 1.00{\small x} & \textbf{1.84}{\small x} & \textbf{2.11}{\small x} & \textbf{1.42}{\small x} & - \\
Mean & \textbf{1.71}{\small x} & 1.10{\small x} & \textbf{2.13}{\small x} & - & 1.02{\small x} & 1.01{\small x} & \textbf{1.68}{\small x} & \textbf{1.29}{\small x} & \textbf{1.71}{\small x} & \textbf{1.25}{\small x} & \textbf{1.03}{\small x} & \textbf{1.00}{\small x} & \underline{2.10}{\small x} & \underline{1.92}{\small x} & 1.05{\small x} & \textbf{1.03}{\small x} & \textbf{1.84}{\small x} & \textbf{2.11}{\small x} & \textbf{1.42}{\small x} & - \\
Merge & \textbf{1.71}{\small x} & \textbf{1.12}{\small x} & 2.12{\small x} & - & 1.02{\small x} & 0.99{\small x} & \textbf{1.68}{\small x} & \textbf{1.29}{\small x} & \textbf{1.71}{\small x} & \textbf{1.25}{\small x} & 1.02{\small x} & 0.99{\small x} & \underline{2.10}{\small x} & \textbf{1.93}{\small x} & \textbf{1.06}{\small x} & 1.01{\small x} & \textbf{1.84}{\small x} & 2.10{\small x} & \textbf{1.42}{\small x} & - \\
Align & \textbf{1.71}{\small x} & 1.11{\small x} & 2.12{\small x} & - & \underline{1.03}{\small x} & 1.02{\small x} & \textbf{1.68}{\small x} & \textbf{1.29}{\small x} & \textbf{1.71}{\small x} & \textbf{1.25}{\small x} & \textbf{1.03}{\small x} & 0.98{\small x} & \underline{2.10}{\small x} & \underline{1.92}{\small x} & \textbf{1.06}{\small x} & \textbf{1.03}{\small x} & \textbf{1.84}{\small x} & 2.10{\small x} & \textbf{1.42}{\small x} & - \\
        \bottomrule
        
    \end{tabular}
    }

    \label{tab:tokens_per_sec}
\end{table*}

\subsubsection{Inference Efficiency} \label{subsec:speedup}

Table \ref{tab:tokens_per_sec} shows the inference efficiency of adapted models across tasks and languages.
We first see that FOCUS tends to exhibit the worst speedups across languages in \textsc{sum}, followed by Merge.
Specifically, its speedups are the worst in 8 out of 10 languages.
These results are consistent with downstream performance and may be due to an inability to effectively generate newly added target tokens, which is essential for improving inference efficiency in a target language.
For \textsc{mt}, we generally see smaller differences between models since its input is mostly in English, and thus, only the output contributes to inference speedups.
Similarly, for classification tasks (\textsc{mc} and \textsc{gmmlu}), the observed speedups are generally similar across models.
This is primarily because these tasks involve processing a long input context, and their output typically requires only a single token, limiting the room for producing the difference between different approaches.

Examining speedups by language and script, most non-Latin script languages (Arabic, Burmese, Hindi, Sinhala, Telugu, and Thai) exhibit substantial inference speedups, reaching up to 4.58x. In contrast, Japanese shows only moderate speedups, up to 1.13x. This is likely due to its prior representation in the Llama2 training corpus, which includes Japanese (0.1\%) and Chinese (0.13\%) data~\cite{Touvron2023Llama2O}, with Chinese sharing common scripts (i.e. \textit{kanji}).
Similarly, German shows negligible speedups across tasks. This can be attributed to its largest non-English language representation (0.17\%) in the Llama2 training corpus, implying its vocabulary is already well-covered. As a Latin script language, German is also inherently well-tokenized by the primarily Latin-script-based vocabulary of Llama2.
This further suggests that expanding the vocabulary of a source model with a small number of target language tokens (i.e. 100 in Table \ref{tab:tokens_per_sec}) does not substantially accelerate inference if the language is already well-represented in the vocabulary of the base model.
Figure \ref{fig:flores} further supports this trend, as Japanese and German both show far less text tokenization overfragmentation for Llama2 compared to Burmese, Greek, Hindi, Sinhala, Telugu, and Thai.
Despite not being explicitly included in the Llama2 training corpus, Swahili shows limited speedups. This might be because its Latin script aligns well with the predominant Latin script pre-training data (91\%), meaning its tokenization is already relatively efficient even without explicit target vocabulary expansion.

\paragraph{Recommendation}
Using Mean or Align can provide better downstream performance on generation tasks and inference speedups across tasks.
Vocabulary expansion contributes well to inference speedups when a target language is not included in pre-training data and is written in non-Latin scripts.\footnote{If inference speedups do not matter, we recommend using CPT-only in general. Since it does not involve parameter updates from scratch, it can lead to more stable task performance in low-resource settings.}

\subsection{Training Strategy} \label{subsec:training_method}

We analyze the effectiveness of each training strategy introduced in \S\ref{sec:training_method}.
Given that LAPT does not typically improve classification performance in low-resource settings, we focus on generation tasks for brevity.\footnote{The corresponding classification results are presented in Table \ref{tab:lapt_strategy_cls} in the Appendix. While some fluctuations are observed, we confirm that no training strategy offers consistent improvements across tasks and languages.}
Due to limited compute, we only use the best-performing Align as the target parameter initialization method and experiment with six languages with the largest speedups (i.e. Arabic, Burmese, Greek, Hindi, Sinhala, and Telugu) in \textsc{sum}.
Table \ref{tab:lapt_strategy} lists the performance of the adapted models.

\setlength{\tabcolsep}{3pt}
\begin{table}[!t]
\caption{
Mean performance of Align models on generation tasks over five runs with Llama2 as source.
\colorbox{green!20}{Green} indicates positive performance change over Source.
}

\begin{center}
\small
\begin{tabular}{llllllllllllll}
\toprule
 & & \multicolumn{2}{c}{\textbf{Arabic}}  
 & \multicolumn{2}{c}{\textbf{Burmese}}
 & \multicolumn{2}{c}{\textbf{Greek}} 
 & \multicolumn{2}{c}{\textbf{Hindi}}
 & \multicolumn{2}{c}{\textbf{Sinhala}}
 & \multicolumn{2}{c}{\textbf{Telugu}}\\

\cmidrule(lr){3-4}
\cmidrule(lr){5-6}
\cmidrule(lr){7-8}
\cmidrule(lr){9-10}
\cmidrule(lr){11-12}
\cmidrule(lr){13-14}

\multicolumn{2}{c}{\textbf{Model}} & \multicolumn{1}{c}{\scriptsize \textsc{mt}} & \multicolumn{1}{c}{\scriptsize \textsc{sum}}
& \multicolumn{1}{c}{\scriptsize \textsc{mt}} & \multicolumn{1}{c}{\scriptsize \textsc{sum}}
& \multicolumn{1}{c}{\scriptsize \textsc{mt}} & \multicolumn{1}{c}{\scriptsize \textsc{sum}}
& \multicolumn{1}{c}{\scriptsize \textsc{mt}} & \multicolumn{1}{c}{\scriptsize \textsc{sum}}
& \multicolumn{1}{c}{\scriptsize \textsc{mt}} & \multicolumn{1}{c}{\scriptsize \textsc{sum}}
& \multicolumn{1}{c}{\scriptsize \textsc{mt}} & \multicolumn{1}{c}{\scriptsize \textsc{sum}}\\

\midrule
\rowcolor{gray!25}

\multicolumn{2}{c}{Source} & .08\textsubscript{{\tiny.03}} & 36\textsubscript{{\tiny0.0}} & .03\textsubscript{{\tiny.00}} & 24\textsubscript{{\tiny0.1}} & .08\textsubscript{{\tiny.03}} & 25\textsubscript{{\tiny0.2}} & .12\textsubscript{{\tiny.03}} & 39\textsubscript{{\tiny0.2}} & .05\textsubscript{{\tiny.00}} & 26\textsubscript{{\tiny0.1}} & .06\textsubscript{{\tiny.00}} & 21\textsubscript{{\tiny0.2}}\\

\multicolumn{2}{c}{CPT-only} & \cellcolor{green!20}.18\textsubscript{{\tiny.00}} & 36\textsubscript{{\tiny0.2}} & \cellcolor{green!20}.11\textsubscript{{\tiny.00}} & 24\textsubscript{{\tiny0.2}} & \cellcolor{green!20}.21\textsubscript{{\tiny.00}} & 25\textsubscript{{\tiny0.3}} & \cellcolor{green!20}.21\textsubscript{{\tiny.00}} & 38\textsubscript{{\tiny0.2}} & \cellcolor{green!20}.07\textsubscript{{\tiny.00}} & 23\textsubscript{{\tiny0.1}} & \cellcolor{green!20}.09\textsubscript{{\tiny.00}} & \cellcolor{green!20}29\textsubscript{{\tiny0.1}}\\
\midrule

\multirow{4}{*}{\rotatebox[origin=c]{90}{\parbox[c]{1.25cm}{\centering \footnotesize \textbf{LoRA}}}} 
& \textsc{clm}+2048 & .06\textsubscript{{\tiny.00}} & 33\textsubscript{{\tiny0.1}} & .04\textsubscript{{\tiny.00}} & 15\textsubscript{{\tiny0.7}} & \cellcolor{green!20}.16\textsubscript{{\tiny.00}} & 17\textsubscript{{\tiny0.2}} & \cellcolor{green!20}.13\textsubscript{{\tiny.00}} & 35\textsubscript{{\tiny0.2}} & \cellcolor{green!20}.08\textsubscript{{\tiny.00}} & \cellcolor{green!20}30\textsubscript{{\tiny0.3}} & .06\textsubscript{{\tiny.00}} & \cellcolor{green!20}27\textsubscript{{\tiny0.2}}\\

& \textsc{mtp}+2048 & \cellcolor{green!20}.11\textsubscript{{\tiny.00}} & 32\textsubscript{{\tiny0.2}} & .03\textsubscript{{\tiny.00}} & 14\textsubscript{{\tiny0.7}} & \cellcolor{green!20}.14\textsubscript{{\tiny.00}} & 20\textsubscript{{\tiny0.2}} & .12\textsubscript{{\tiny.00}} & 34\textsubscript{{\tiny0.2}} & \cellcolor{green!20}.08\textsubscript{{\tiny.00}} & \cellcolor{green!20}29\textsubscript{{\tiny0.1}} & .04\textsubscript{{\tiny.00}} & \cellcolor{green!20}25\textsubscript{{\tiny0.1}}\\

& \textsc{clm}+512 & \cellcolor{green!20}.12\textsubscript{{\tiny.00}} & 33\textsubscript{{\tiny0.1}} & \cellcolor{green!20}.12\textsubscript{{\tiny.00}} & 23\textsubscript{{\tiny0.1}} & \cellcolor{green!20}.23\textsubscript{{\tiny.00}} & 19\textsubscript{{\tiny0.4}} & \cellcolor{green!20}.17\textsubscript{{\tiny.00}} & 35\textsubscript{{\tiny0.1}} & \cellcolor{green!20}.11\textsubscript{{\tiny.00}} & \cellcolor{green!20}32\textsubscript{{\tiny0.1}} & \cellcolor{green!20}.10\textsubscript{{\tiny.00}} & \cellcolor{green!20}28\textsubscript{{\tiny0.1}}\\

& \textsc{mtp}+512 & \cellcolor{green!20}.14\textsubscript{{\tiny.00}} & 33\textsubscript{{\tiny0.1}} & \cellcolor{green!20}.12\textsubscript{{\tiny.00}} & 23\textsubscript{{\tiny0.1}} & \cellcolor{green!20}.21\textsubscript{{\tiny.00}} & 21\textsubscript{{\tiny0.3}} & \cellcolor{green!20}.15\textsubscript{{\tiny.00}} & 35\textsubscript{{\tiny0.1}} & \cellcolor{green!20}.11\textsubscript{{\tiny.00}} & \cellcolor{green!20}30\textsubscript{{\tiny0.3}} & \cellcolor{green!20}.09\textsubscript{{\tiny.00}} & \cellcolor{green!20}28\textsubscript{{\tiny0.1}}\\

\midrule

\multirow{4}{*}{\rotatebox[origin=c]{90}{\parbox[c]{1.25cm}{\centering \footnotesize \textbf{2-stage}}}} 
& \textsc{clm}+2048 & \cellcolor{green!20}.09\textsubscript{{\tiny.00}} & 33\textsubscript{{\tiny0.1}} & \cellcolor{green!20}.04\textsubscript{{\tiny.00}} & 17\textsubscript{{\tiny0.2}} & \cellcolor{green!20}.13\textsubscript{{\tiny.00}} & 21\textsubscript{{\tiny0.1}} & \cellcolor{green!20}.13\textsubscript{{\tiny.00}} & 36\textsubscript{{\tiny0.1}} & \cellcolor{green!20}.09\textsubscript{{\tiny.00}} & \cellcolor{green!20}30\textsubscript{{\tiny0.4}} & .05\textsubscript{{\tiny.00}} & \cellcolor{green!20}28\textsubscript{{\tiny0.1}}\\

& \textsc{mtp}+2048 & \cellcolor{green!20}.13\textsubscript{{\tiny.00}} & 33\textsubscript{{\tiny0.1}} & .02\textsubscript{{\tiny.00}} & 17\textsubscript{{\tiny0.3}} & \cellcolor{green!20}.19\textsubscript{{\tiny.00}} & 21\textsubscript{{\tiny0.1}} & .11\textsubscript{{\tiny.00}} & 36\textsubscript{{\tiny0.1}} & \cellcolor{green!20}.08\textsubscript{{\tiny.00}} & \cellcolor{green!20}29\textsubscript{{\tiny0.2}} & .03\textsubscript{{\tiny.00}} & \cellcolor{green!20}25\textsubscript{{\tiny0.2}}\\

& \textsc{clm}+512 & \cellcolor{green!20}.12\textsubscript{{\tiny.01}} & 33\textsubscript{{\tiny0.1}} & \cellcolor{green!20}.07\textsubscript{{\tiny.00}} & 13\textsubscript{{\tiny0.5}} & \cellcolor{green!20}.22\textsubscript{{\tiny.00}} & 22\textsubscript{{\tiny0.3}} & \cellcolor{green!20}.16\textsubscript{{\tiny.00}} & 35\textsubscript{{\tiny0.1}} & \cellcolor{green!20}.10\textsubscript{{\tiny.00}} & \cellcolor{green!20}31\textsubscript{{\tiny0.3}} & \cellcolor{green!20}.07\textsubscript{{\tiny.00}} & \cellcolor{green!20}28\textsubscript{{\tiny0.1}}\\

& \textsc{mtp}+512 & \cellcolor{green!20}.11\textsubscript{{\tiny.00}} & 33\textsubscript{{\tiny0.2}} & \cellcolor{green!20}.07\textsubscript{{\tiny.00}} & 17\textsubscript{{\tiny0.2}} & \cellcolor{green!20}.21\textsubscript{{\tiny.00}} & 22\textsubscript{{\tiny0.1}} & \cellcolor{green!20}.16\textsubscript{{\tiny.00}} & 35\textsubscript{{\tiny0.1}} & \cellcolor{green!20}.08\textsubscript{{\tiny.00}} & \cellcolor{green!20}31\textsubscript{{\tiny0.3}} & \cellcolor{green!20}.07\textsubscript{{\tiny.00}} & \cellcolor{green!20}28\textsubscript{{\tiny0.1}}\\

\midrule

\multirow{4}{*}{\rotatebox[origin=c]{90}{\parbox[c]{1.5cm}{\centering \footnotesize \textbf{2x2 LS}}}} 
& \textsc{clm}+2048 & \cellcolor{green!20}.11\textsubscript{{\tiny.01}} & 33\textsubscript{{\tiny0.1}} & \cellcolor{green!20}.14\textsubscript{{\tiny.00}} & \cellcolor{green!20}25\textsubscript{{\tiny0.1}} & \cellcolor{green!20}.24\textsubscript{{\tiny.00}} & 22\textsubscript{{\tiny0.3}} & \cellcolor{green!20}.15\textsubscript{{\tiny.00}} & 36\textsubscript{{\tiny0.2}} & \cellcolor{green!20}.11\textsubscript{{\tiny.00}} & \cellcolor{green!20}32\textsubscript{{\tiny0.2}} & \cellcolor{green!20}.11\textsubscript{{\tiny.00}} & \cellcolor{green!20}29\textsubscript{{\tiny0.1}}\\

& \textsc{mtp}+2048 & \cellcolor{green!20}.15\textsubscript{{\tiny.01}} & 33\textsubscript{{\tiny0.1}} & \cellcolor{green!20}.11\textsubscript{{\tiny.00}} & 23\textsubscript{{\tiny0.2}} & \cellcolor{green!20}.23\textsubscript{{\tiny.00}} & 23\textsubscript{{\tiny0.2}} & .12\textsubscript{{\tiny.00}} & 36\textsubscript{{\tiny0.0}} & \cellcolor{green!20}.12\textsubscript{{\tiny.00}} & \cellcolor{green!20}32\textsubscript{{\tiny0.1}} & \cellcolor{green!20}.12\textsubscript{{\tiny.00}} & \cellcolor{green!20}28\textsubscript{{\tiny0.2}}\\

& \textsc{clm}+512 & \cellcolor{green!20}.11\textsubscript{{\tiny.00}} & 33\textsubscript{{\tiny0.1}} & \cellcolor{green!20}.15\textsubscript{{\tiny.00}} & \cellcolor{green!20}26\textsubscript{{\tiny0.2}} & \cellcolor{green!20}.25\textsubscript{{\tiny.00}} & 23\textsubscript{{\tiny0.3}} & \cellcolor{green!20}.19\textsubscript{{\tiny.00}} & 36\textsubscript{{\tiny0.1}} & \cellcolor{green!20}.11\textsubscript{{\tiny.00}} & \cellcolor{green!20}33\textsubscript{{\tiny0.2}} & \cellcolor{green!20}.13\textsubscript{{\tiny.00}} & \cellcolor{green!20}30\textsubscript{{\tiny0.1}}\\

& \textsc{mtp}+512 & \cellcolor{green!20}.17\textsubscript{{\tiny.00}} & 33\textsubscript{{\tiny0.2}} & \cellcolor{green!20}.16\textsubscript{{\tiny.00}} & \cellcolor{green!20}26\textsubscript{{\tiny0.2}} & \cellcolor{green!20}.24\textsubscript{{\tiny.00}} & 23\textsubscript{{\tiny0.2}} & \cellcolor{green!20}.15\textsubscript{{\tiny.00}} & 36\textsubscript{{\tiny0.1}} & \cellcolor{green!20}.13\textsubscript{{\tiny.00}} & \cellcolor{green!20}33\textsubscript{{\tiny0.3}} & \cellcolor{green!20}.12\textsubscript{{\tiny.00}} & \cellcolor{green!20}29\textsubscript{{\tiny0.1}}\\

\bottomrule
\end{tabular}%
\label{tab:lapt_strategy}
\end{center}
\end{table}

Looking at the results by training procedure, we observe that 2x2 LS improves performance across tasks and languages.
Gains range from 1 point (Hindi \textsc{sum}) to 10 points (Burmese \textsc{mt}) compared to LoRA with \textsc{clm}+2048 (our default approach in \S\ref{subsec:vocab_init}).
The sole exception is Arabic \textsc{sum}, where performance remains the same as the baseline.
Notably, 2x2 LS also helps to partially mitigate performance drops observed in \textsc{sum} for languages like Burmese and Greek.
We hypothesize that 2x2 LS can reduce the risk of underfitting by focusing on calibrating only the parts closely related to the encoding and decoding of the target language.
This suggests that the frequently used full LoRA approach in high-resource settings is not the best training approach in low-resource settings.
Although 2-stage does not consistently provide gains over LoRA, especially for \textsc{mt}, it does not lead to substantial performance degradation either, with a maximum drop of 3 points for Greek \textsc{mt}, compared to the default LoRA with \textsc{clm}+2048.

Next, we examine the effectiveness of \textsc{mtp}. Although it does not consistently improve performance over \textsc{clm}, it notably boosts Arabic \textsc{mt} performance across models, particularly when used with 2x2 LS. Similar to the 2-stage approach, any performance degradation observed with \textsc{mtp} is minor, staying within 2 points. The only exceptions across all cases are Burmese and Hindi \textsc{mt} with 2x2 LS, and Telugu \textsc{sum} with 2-stage, where we observe moderate drops of 3 to 4 points.
Based on these results, while \textsc{mtp} does not offer universal gains, its consistent benefit for a specific language like Arabic \textsc{mt} and its generally stable performance (i.e. avoiding substantial drops) across other tasks make it a valuable strategy, particularly when used with 2x2 LS.\footnote{It is worth noting that \textsc{mtp} typically offers an additional benefit of improving inference efficiency through self-speculative decoding~\cite{pmlr-v235-gloeckle24a}, which can be a huge advantage in practical deployments, though this aspect is not the primary focus of our analysis and beyond the scope of this article.}

Finally, using a short sequence length (512) works well across models and languages, showing improvements of up to 9 points (LoRA+\textsc{mtp} in Burmese and 2-stage+\textsc{clm} in Greek \textsc{mt}), confirming our hypothesis (\S\ref{sec:training_method}) that increasing the number of model updates can help avoid underfitting.

\paragraph{Recommendation}
2x2 LS generally leads to performance improvements. The short training sequence length of 512 can also aid low-resource settings. While \textsc{mtp} does not always offer improvements, its notable boost for a certain language and its overall stable performance, with no substantial degradation elsewhere, make it a viable and beneficial strategy for adaptation in low-resource environments.

\subsection{Experiments with Other Source LLMs} \label{subsec:llama3}

We investigate whether models other than Llama2 adapted based on our recommendations in \S\ref{subsec:vocab_init} and \S\ref{subsec:training_method} benefit from inference speedups while maintaining competitive performance to their base models.
We use Llama3 and Gemma2 as source.
We apply the 2x2 LS+\textsc{mtp}+512 training strategy and initialize models with either Mean or Align along with Random as a baseline.
Due to resource constraints, we only experiment with Burmese, Sinhala, and Telugu in the remainder of the article, as they are the worst fragmented languages of our target languages (Figure \ref{fig:flores}).
Table \ref{tab:llama3_result} shows the performance of adapted models using Llama3 and Gemma2 as source.

\setlength{\tabcolsep}{4pt}
\renewcommand*{\arraystretch}{1.0}
\begin{table}[!t]
\centering
\small
\caption{Mean performance and inference speedup with Llama3 and Gemma2 as source. \colorbox{green!20}{Green} indicates positive performance change over Source. The speedup ratio corresponds to Align.
}

\begin{center}
\resizebox{\linewidth}{!}{%
\begin{tabular}{llllllllllllll}
\toprule
 & & \multicolumn{4}{c}{\textbf{Burmese}}  
 & \multicolumn{4}{c}{\textbf{Sinhala}} 
 & \multicolumn{4}{c}{\textbf{Telugu}}\\

\cmidrule(lr){3-6}
\cmidrule(lr){7-10}
\cmidrule(lr){11-14}

\multicolumn{2}{l}{\textbf{Llama3}} & \multicolumn{1}{c}{\scriptsize \textsc{mt}} & \multicolumn{1}{c}{\scriptsize \textsc{sum}}
& \multicolumn{1}{c}{\scriptsize \textsc{mc}} & \multicolumn{1}{c}{\scriptsize \textsc{gmmlu}}
& \multicolumn{1}{c}{\scriptsize \textsc{mt}} & \multicolumn{1}{c}{\scriptsize \textsc{sum}}
& \multicolumn{1}{c}{\scriptsize \textsc{mc}} & \multicolumn{1}{c}{\scriptsize \textsc{gmmlu}}
& \multicolumn{1}{c}{\scriptsize \textsc{mt}} & \multicolumn{1}{c}{\scriptsize \textsc{sum}}
& \multicolumn{1}{c}{\scriptsize \textsc{mc}} & \multicolumn{1}{c}{\scriptsize \textsc{gmmlu}}\\

\midrule
\rowcolor{gray!25}
& Source & .09\textsubscript{{\tiny.00}} & 28\textsubscript{{\tiny0.0}} & .41 & -
& .19\textsubscript{{\tiny.00}} & 30\textsubscript{{\tiny0.0}} & .50 & .37
& .24\textsubscript{{\tiny.00}} & 29\textsubscript{{\tiny0.0}} & .51 & .40\\

& CPT-only & \cellcolor{green!20}.17\textsubscript{{\tiny.00}} & 28\textsubscript{{\tiny0.1}} & .31 & -
& \cellcolor{green!20}.21\textsubscript{{\tiny.00}} & \cellcolor{green!20}31\textsubscript{{\tiny0.1}} & .47 & .36
& \cellcolor{green!20}.30\textsubscript{{\tiny.00}} & 29\textsubscript{{\tiny0.1}} & .42 & .38\\

\midrule
\multirow{3}{*}{\rotatebox[origin=c]{90}{\parbox[c]{1cm}{\centering \tiny \textbf{+Speedup}}}} 
& Random & \cellcolor{green!20}.19\textsubscript{{\tiny.00}} & 23\textsubscript{{\tiny0.2}} & .28 & -
& .14\textsubscript{{\tiny.00}} & \cellcolor{green!20}31\textsubscript{{\tiny0.2}} & .23 & .25 
& .13\textsubscript{{\tiny.00}} & 29\textsubscript{{\tiny0.1}} & .27 & .26\\

 & Mean & \cellcolor{green!20}.19\textsubscript{{\tiny.00}} & 23\textsubscript{{\tiny0.2}} & .25 & -
 & .12\textsubscript{{\tiny.00}} & \cellcolor{green!20}32\textsubscript{{\tiny0.1}} & .23 & .23
 & .12\textsubscript{{\tiny.00}} & 29\textsubscript{{\tiny0.2}} & .26 & .25\\

& Align & \cellcolor{green!20}.21\textsubscript{{\tiny.00}} & 28\textsubscript{{\tiny0.3}} & .28 & - 
& \cellcolor{green!20}.22\textsubscript{{\tiny.00}} & \cellcolor{green!20}33\textsubscript{{\tiny0.1}} & .30 & .26
& \cellcolor{green!20}.31\textsubscript{{\tiny.00}} & \cellcolor{green!20}31\textsubscript{{\tiny0.1}} & .38 & .31\\

\midrule
\multicolumn{2}{l}{\textbf{Speedup}} & 2.60{\small x} & 3.52{\small x} & 2.30{\small x} & -
& 2.36{\small x} & 3.19{\small x} & 1.98{\small x} & 1.87{\small x}
& 2.15{\small x} & 3.55{\small x} & 1.96{\small x} & 1.85{\small x} \\

\bottomrule

\\
\toprule
 & & \multicolumn{4}{c}{\textbf{Burmese}}  
 & \multicolumn{4}{c}{\textbf{Sinhala}} 
 & \multicolumn{4}{c}{\textbf{Telugu}}\\

\cmidrule(lr){3-6}
\cmidrule(lr){7-10}
\cmidrule(lr){11-14}

\multicolumn{2}{l}{\textbf{Gemma2}} & \multicolumn{1}{c}{\scriptsize \textsc{mt}} & \multicolumn{1}{c}{\scriptsize \textsc{sum}}
& \multicolumn{1}{c}{\scriptsize \textsc{mc}} & \multicolumn{1}{c}{\scriptsize \textsc{gmmlu}}
& \multicolumn{1}{c}{\scriptsize \textsc{mt}} & \multicolumn{1}{c}{\scriptsize \textsc{sum}}
& \multicolumn{1}{c}{\scriptsize \textsc{mc}} & \multicolumn{1}{c}{\scriptsize \textsc{gmmlu}}
& \multicolumn{1}{c}{\scriptsize \textsc{mt}} & \multicolumn{1}{c}{\scriptsize \textsc{sum}}
& \multicolumn{1}{c}{\scriptsize \textsc{mc}} & \multicolumn{1}{c}{\scriptsize \textsc{gmmlu}}\\

\midrule
\rowcolor{gray!25}

& Source & .21\textsubscript{{\tiny.01}} & 34\textsubscript{{\tiny0.1}} & .67 & -
& .19\textsubscript{{\tiny.00}} & 34\textsubscript{{\tiny0.2}} & .71 & .50
& .31\textsubscript{{\tiny.01}} & 30\textsubscript{{\tiny0.2}} & .74 & .57\\

& CPT-only & \cellcolor{green!20}.28\textsubscript{{\tiny.00}} & 28\textsubscript{{\tiny0.3}} & .61 & -
& \cellcolor{green!20}.29\textsubscript{{\tiny.00}} & 34\textsubscript{{\tiny0.1}} & .69 & .48
& \cellcolor{green!20}.43\textsubscript{{\tiny.00}} & 29\textsubscript{{\tiny0.1}} & .71 & .56\\

\midrule
\multirow{3}{*}{\rotatebox[origin=c]{90}{\parbox[c]{1cm}{\centering \tiny \textbf{+Speedup}}}} 
& Random & \cellcolor{green!20}.24\textsubscript{{\tiny.00}} & 26\textsubscript{{\tiny0.3}} & .46 & -
& \cellcolor{green!20}.26\textsubscript{{\tiny.00}} & 33\textsubscript{{\tiny0.1}} & .63 & .43
& \cellcolor{green!20}.35\textsubscript{{\tiny.00}} & 28\textsubscript{{\tiny0.2}}& .65 & .47\\

& Mean & \cellcolor{green!20}.25\textsubscript{{\tiny.00}} & 25\textsubscript{{\tiny0.3}} & .49 & -
& \cellcolor{green!20}.27\textsubscript{{\tiny.00}} & 32\textsubscript{{\tiny0.2}} & .63 & .42
& \cellcolor{green!20}.33\textsubscript{{\tiny.00}} & 28\textsubscript{{\tiny0.1}} & .60 & .43\\

& Align & \cellcolor{green!20}.25\textsubscript{{\tiny.00}} & 25\textsubscript{{\tiny0.2}} & .32 & - 
& \cellcolor{green!20}.28\textsubscript{{\tiny.00}} & 32\textsubscript{{\tiny0.1}} & .63 & .42
& \cellcolor{green!20}.32\textsubscript{{\tiny.00}} & 29\textsubscript{{\tiny0.1}} & .59 & .43\\

\midrule
\multicolumn{2}{l}{\textbf{Speedup}} & 1.52{\small x} & 1.57{\small x} & 1.51{\small x} & - & 1.26{\small x} & 1.38{\small x} & 1.31{\small x} & 1.20{\small x} & 1.07{\small x} & 1.10{\small x} & 1.06{\small x} & 1.13{\small x} \\

\bottomrule
\end{tabular}%
}
\label{tab:llama3_result}
\end{center}
\end{table}

\setlength{\tabcolsep}{3pt}
\begin{table}[t!]
    \caption{Perplexity on language-specific held-out dataset using Llama3 and Gemma2 as source. Results are not comparable between models with \colorbox{gray!25}{gray} and others due to their difference in vocabulary.}
    \begin{minipage}{0.45\textwidth}
    \small
    \centering
    \begin{center} 
    \begin{tabular}{llll}
    \multicolumn{4}{c}{(a) Llama3}\\
    \toprule
    \textbf{Model} & \textbf{my} & \textbf{si} & \textbf{te}\\
    
    \midrule
    \rowcolor{gray!25}
    Source & 3.7 & 3.5 & 3.0\\
    \rowcolor{gray!25}
    CPT-only & 1.9 & 2.0 & 1.8\\
    \midrule
    Random & 6.9 & 7.1 & 6.1\\
    Mean & 6.7 & 6.8 & 5.8\\
    Align & 6.0 & 5.5 & 4.6\\
    \bottomrule
    \end{tabular}%
    \end{center}
\end{minipage}
\hfill
\begin{minipage}{0.45\textwidth}
    \small
    \centering
    \begin{center} 
    \begin{tabular}{llll}
    \multicolumn{4}{c}{(b) Gemma2}\\
    \toprule
    \textbf{Model} & \textbf{my} & \textbf{si} & \textbf{te}\\
    
    \midrule
    \rowcolor{gray!25}
    Source & 48.0 & 59.9 & 51.9\\
    \rowcolor{gray!25}
    CPT-only & 4.6 & 4.3 & 5.2\\
    \midrule
    Random & 9.0 & 6.9 & 6.7\\
    Mean & 9.0 & 6.9 & 6.6\\
    Align & 9.0 & 6.9 & 6.6\\
    \bottomrule
    \end{tabular}%
    \end{center}
\end{minipage}
\label{tab:ppl_result_llama3}
\end{table}

Overall, models adapted with Align generally either perform competitively or outperform their respective Llama3 and Gemma2 base models (Source) on both generation tasks. The only exception is Burmese \textsc{sum} with Gemma2, which shows a substantial 9-point drop.
In contrast, models adapted with Mean, along with Random, often underperform Align when using Llama3 as the base model.
Specifically, they underperform Align in Burmese \textsc{sum}, Sinhala and Telugu \textsc{mt} tasks, and across almost all classification tasks. This underperformance is attributed to underfitting, as evidenced by their higher perplexities compared to Align (see Table \ref{tab:ppl_result_llama3}).

A consistent challenge across vocabulary expansion approaches, including Align, is substantial performance degradation on classification tasks. This degradation ranges from an 8-point drop (Sinhala \textsc{mc} and \textsc{gmmlu} with Gemma2) to a significant 35-point drop (Burmese \textsc{mc} with Gemma2) for the best-performing Align.
This issue is not unique to these approaches, as CPT-only also frequently exhibits moderate to substantial degradation (3 to 10 points) on classification tasks across both models. 
This further highlights the inherent challenges of LAPT for classification tasks in low-resource settings, as discussed in \S\ref{subsec:performance}. We later showcase in \S\ref{subsec:elchat} that these substantial drops can be largely mitigated using a post-hoc, training-free method, achieving performance within 3 and 5 points of the Gemma2 base model for \textsc{mc} and \textsc{gmmlu}, respectively.

\begin{figure}[t]
\begin{center}
\includegraphics[width=0.65\columnwidth]{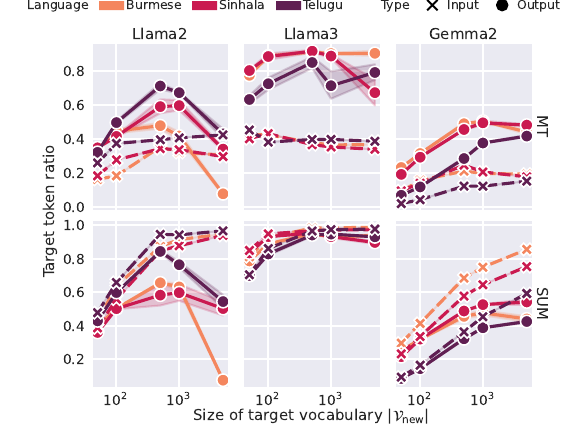}
\caption{
Average target token ratio in input ($\times$) and output ($\bullet$) with respect to $|\mathcal{V}_\text{new}|$ across models, languages, and tasks.
}
\label{fig:target_token_ratio}
\end{center}
\end{figure}

Beyond downstream performance, a core advantage of vocabulary expansion lies in inference efficiency. Align models, for instance, consistently exhibit inference speedups, reaching up to 3.52x for Llama3 and 1.57x for Gemma2.\footnote{For context, speedups in the range of 1.3x to 3x are commonly reported for speculative decoding techniques~\cite{pmlr-v262-agrawal24a,timor2025accelerating,huang2025specdecboostingspeculativedecoding}. Thus, the observed speedups generally represent a meaningful gain in inference efficiency for LLMs.} These results confirm the versatility of our approaches in optimizing inference speed across different base model architectures.
However, we note a relatively moderate inference speedup of up to 1.13x for Telugu with Gemma2, despite its higher text fragmentation rate than Sinhala (Figure \ref{fig:flores}). This can be attributed to lower target token ratios across tasks for both input and output contexts (Figure \ref{fig:target_token_ratio}), suggesting an underutilization of newly added tokens and highlighting potential avenues for improvement in the selection of new tokens for specific languages.

\setlength{\tabcolsep}{3pt}
\begin{table}[t!]
\caption{Mean \textsc{sum} performance in ROUGE-L and BLEURT over five runs. \colorbox{green!20}{Green} indicates positive performance change over Source.
}
\begin{center}
\small
\begin{tabular}{llcccccc}
\toprule
 & & \multicolumn{2}{c}{\textbf{Burmese}}  
 & \multicolumn{2}{c}{\textbf{Sinhala}} 
 & \multicolumn{2}{c}{\textbf{Telugu}}\\

\multicolumn{2}{l}{\textbf{Llama2}} & \multicolumn{1}{c}{\scriptsize ROUGE-L} & \multicolumn{1}{c}{\scriptsize BLEURT}
& \multicolumn{1}{c}{\scriptsize ROUGE-L} & \multicolumn{1}{c}{\scriptsize BLEURT}
& \multicolumn{1}{c}{\scriptsize ROUGE-L} & \multicolumn{1}{c}{\scriptsize BLEURT}\\

\midrule
\rowcolor{gray!25}
& Source  
& 24\textsubscript{{\tiny0.1}} & .03\textsubscript{{\tiny.00}} & 
26\textsubscript{{\tiny0.1}} & .19\textsubscript{{\tiny.00}} & 
21\textsubscript{{\tiny0.2}} & .13\textsubscript{{\tiny.00}}\\

& CPT-only
& 24\textsubscript{{\tiny0.3}}  & \cellcolor{green!20}.04\textsubscript{{\tiny.00}} & 
23\textsubscript{{\tiny.0.1}} & \cellcolor{green!20}.25\textsubscript{{\tiny.00}} & 
\cellcolor{green!20}29\textsubscript{{\tiny0.1}} & \cellcolor{green!20}.22\textsubscript{{\tiny.00}}\\

\midrule
\multirow{3}{*}{\rotatebox[origin=c]{90}{\parbox[c]{1cm}{\centering \tiny \textbf{+Speedup}}}} 
& Random 
& 24\textsubscript{{\tiny0.1}}  & \cellcolor{green!20}.07\textsubscript{{\tiny.00}} 
& \cellcolor{green!20}32\textsubscript{{\tiny0.2}} & \cellcolor{green!20}.26\textsubscript{{\tiny.00}}
& \cellcolor{green!20}29\textsubscript{{\tiny0.1}} & \cellcolor{green!20}.23\textsubscript{{\tiny.00}}\\

& Mean  
& \cellcolor{green!20}25\textsubscript{{\tiny0.1}}  & \cellcolor{green!20}.06\textsubscript{{\tiny.00}} & \cellcolor{green!20}32\textsubscript{{\tiny0.1}} &\cellcolor{green!20}.25\textsubscript{{\tiny.00}} & \cellcolor{green!20}
29\textsubscript{{\tiny0.1}} & \cellcolor{green!20}.23\textsubscript{{\tiny.01}}\\

& Align  
& \cellcolor{green!20}26\textsubscript{{\tiny0.2}} & \cellcolor{green!20}.08\textsubscript{{\tiny.01}} 
& \cellcolor{green!20}33\textsubscript{{\tiny0.3}}& \cellcolor{green!20}.26\textsubscript{{\tiny.01}} 
& \cellcolor{green!20}29\textsubscript{{\tiny0.1}} & \cellcolor{green!20}.24\textsubscript{{\tiny.00}}\\

\bottomrule

\\
\multicolumn{2}{l}{\textbf{Llama3}}\\

\midrule
\rowcolor{gray!25}
& Source  
& 28\textsubscript{{\tiny.00}} & .06\textsubscript{{\tiny.00}} & 
30\textsubscript{{\tiny.00}} & .24\textsubscript{{\tiny.00}} & 
29\textsubscript{{\tiny.00}} & .22\textsubscript{{\tiny.00}}\\

& CPT-only 
& 28\textsubscript{{\tiny0.1}} & .06\textsubscript{{\tiny.00}} & 
\cellcolor{green!20}31\textsubscript{{\tiny0.1}} & \cellcolor{green!20}.25\textsubscript{{\tiny.00}} & 
\cellcolor{green!20}30\textsubscript{{\tiny0.1}} & \cellcolor{green!20}.29\textsubscript{{\tiny.00}}\\

\midrule
\multirow{3}{*}{\rotatebox[origin=c]{90}{\parbox[c]{1cm}{\centering \tiny \textbf{+Speedup}}}} 
& Random 
& 23\textsubscript{{\tiny0.2}}  & .05\textsubscript{{\tiny.00}} 
& \cellcolor{green!20}31\textsubscript{{\tiny0.2}}  & \cellcolor{green!20}.27\textsubscript{{\tiny.00}} 
& 29\textsubscript{{\tiny0.1}} & \cellcolor{green!20}.24\textsubscript{{\tiny.01}}\\

& Mean  
& 23\textsubscript{{\tiny0.2}}  & .05\textsubscript{{\tiny.00}} & 
\cellcolor{green!20}32\textsubscript{{\tiny0.1}} & \cellcolor{green!20}.29\textsubscript{{\tiny.01}} & 
29\textsubscript{{\tiny0.2}}& \cellcolor{green!20}.26\textsubscript{{\tiny.00}}\\

& Align  
& 28\textsubscript{{\tiny0.3}} & \cellcolor{green!20}.07\textsubscript{{\tiny.00}} 
& \cellcolor{green!20}33\textsubscript{{\tiny0.1}} & \cellcolor{green!20}.31\textsubscript{{\tiny.00}} 
& \cellcolor{green!20}31\textsubscript{{\tiny0.1}} & \cellcolor{green!20}.28\textsubscript{{\tiny.00}}\\

\bottomrule
\\
\multicolumn{2}{l}{\textbf{Gemma2}}\\
\midrule
\rowcolor{gray!25}
& Source  
& 34\textsubscript{{\tiny0.1}} & .10\textsubscript{{\tiny.00}} 
& 34\textsubscript{{\tiny0.2}} & .32\textsubscript{{\tiny.00}} 
& 30\textsubscript{{\tiny0.2}} & .30\textsubscript{{\tiny.00}}\\

& CPT-only 
& 28\textsubscript{{\tiny0.3}} & .10\textsubscript{{\tiny.00}} 
& 34\textsubscript{{\tiny0.1}} & .32\textsubscript{{\tiny.01}} 
& 29\textsubscript{{\tiny0.1}} & .26\textsubscript{{\tiny.00}}\\

\midrule
\multirow{3}{*}{\rotatebox[origin=c]{90}{\parbox[c]{1cm}{\centering \tiny \textbf{+Speedup}}}} 
& Random 
& 26\textsubscript{{\tiny0.3}} & .10\textsubscript{{\tiny.00}} 
& 33\textsubscript{{\tiny0.1}} & \cellcolor{green!20}.34\textsubscript{{\tiny.01}} & 
28\textsubscript{{\tiny0.2}} & .26\textsubscript{{\tiny.00}}\\

& Mean 
& 25\textsubscript{{\tiny0.3}} & .09\textsubscript{{\tiny.00}} 
& 32\textsubscript{{\tiny0.2}} & \cellcolor{green!20}.33\textsubscript{{\tiny.00}} 
& 28\textsubscript{{\tiny0.1}} & .27\textsubscript{{\tiny.00}}\\

& Align 
& 25\textsubscript{{\tiny0.2}} & .10\textsubscript{{\tiny.00}} 
& 32\textsubscript{{\tiny0.1}} & \cellcolor{green!20}.33\textsubscript{{\tiny.01}} 
& 29\textsubscript{{\tiny0.1}} & .28\textsubscript{{\tiny.00}}\\

\bottomrule
\end{tabular}%
\label{tab:llama3_result_appendix}
\end{center}
\end{table}

Finally, examining the results by source model and generation task, we observe that adapted Llama3 and Gemma2 models show substantially better \textsc{mt} performance than those with Llama2 despite their similar model size.
In particular, the Telugu-adapted models initialized with Align with Llama3 and Gemma2 as source obtain 19 and 20 points better performance than those Llama2 counterparts.
This trend suggests a successful cross-lingual transfer of the generative capabilities of the base models.
However, \textsc{sum} does not seem to follow the trend, i.e. the adapted models with Llama3 and Gemma2 do not outperform those with Llama2 when evaluated with ROUGE-L.
In fact, their semantic-level performances measured by BLEURT~\cite{sellam-etal-2020-bleurt} improve up to 7 points from those with Llama2 (see Table \ref{tab:llama3_result_appendix}).
Thus, adapted models actually enjoy their base model capabilities to improve performance at the semantic level rather than the surface level measured by ROUGE-L.

\subsection{Target Vocabulary Size} \label{subsec:vocab_size}

The extent to which a target language benefits from inference speedups varies across languages, models, and tasks, as observed from Tables \ref{tab:tokens_per_sec} and \ref{tab:llama3_result}.
Although larger $|\mathcal{V}_\text{new}|$ can lead to faster inference, it comes with the risk of underfitting in low-resource settings.\footnote{Given a small and limited $\mathcal{D}$, the larger $|\mathcal{V}_\text{new}|$, the less fragmentation with respect to $\mathcal{D}$, the fewer the number of training tokens, and therefore the fewer model updates.}
To help select an optimal $|\mathcal{V}_\text{new}|$ for obtaining substantial speedups while retaining competitive performance to Source, we conduct a cost-benefit analysis with respect to $|\mathcal{V}_\text{new}|$.
We experiment with $|\mathcal{V}_\text{new}| = \{50, 100, 500, 1\text{K}, 5\text{K}\}$ while previous studies set $|\mathcal{V}_\text{new}|$ to over 10K, e.g. 17,953 for Chinese Llama.
For comprehensiveness, we employ all three source models and follow the same setup as in \S\ref{subsec:llama3}.\footnote{Table \ref{tab:tokenization} in the Appendix provides a qualitative example of how tokenization in Burmese changes with respect to $|\mathcal{V}_\text{new}|$ using Llama2 as source.}

\begin{figure}[!t]
\centering
\begin{minipage}{0.49\textwidth}
    \centering
    \includegraphics[width=\textwidth]{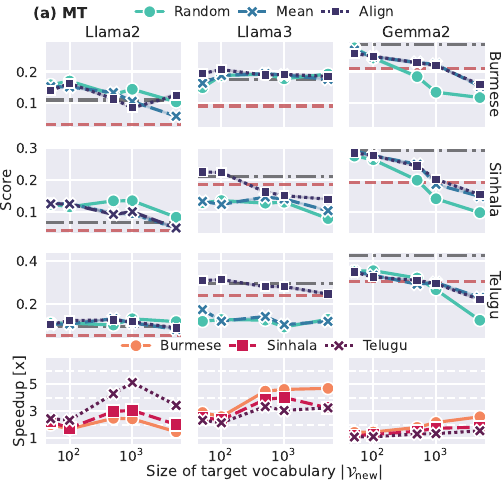}
\end{minipage}
\hfill
\begin{minipage}{0.49\textwidth}
    \centering
    \includegraphics[width=\textwidth]{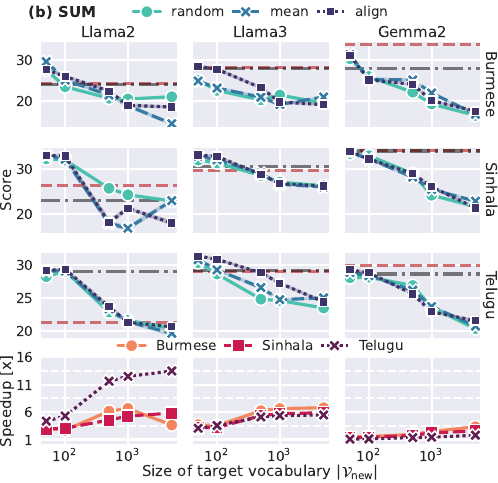}
\end{minipage}
\caption{
Downstream performance and inference speedup in (a) \textsc{mt} and (b) \textsc{sum} across different $|\mathcal{V}_\text{new}|$.
\textcolor{red}{Red} and \textcolor{gray}{gray} dotted lines denote Source and CPT-only.
}
\label{fig:performance_by_vocab}
\end{figure}

\subsubsection{Generation Tasks}
Figure \ref{fig:performance_by_vocab} shows the performance changes and corresponding inference speedups for different $|\mathcal{V}_\text{new}|$ on generation tasks.
We first observe that the larger $|\mathcal{V}_\text{new}|$, the worse the task performance across different source models, languages, and tasks in general.
Notably, \textsc{sum} appears to be less robust to the changes in $|\mathcal{V}_\text{new}|$ than \textsc{mt} across different models and languages.
Adapted models underperform Source even at $|\mathcal{V}_\text{new}| = 500$ in the majority of the cases, while they, especially when initialized with Align, still outperform or rival Source in \textsc{mt} in almost all the cases.
This difference can be due to the difficulty of each task as discussed in \S\ref{subsec:vocab_init}, i.e. \textsc{sum} requires more $\mathcal{D}$ than \textsc{mt} to perform well.
These suggest that $|\mathcal{V}_\text{new}|$ should be set smaller (e.g. $|\mathcal{V}_\text{new}| = 100$ in our setup) for tasks like \textsc{sum} to be competitive with Source in task performance.

Next, we observe that the larger $|\mathcal{V}_\text{new}|$, the faster the inference up to around $|\mathcal{V}_\text{new}|=\text{1K}$ in all the cases.
After that point, the inference speedups tend to plateau or sometimes deteriorate.
This can be partially due to underfitting, which hinders a model from effectively using $\mathcal{V}_\text{new}$.
Indeed, the target token ratio in output at $|\mathcal{V}_\text{new}| = \text{5K}$ drops by 59.7\% (Burmese), 9.5\% (Sinhala), and 30.0\% (Telugu) from its peak at $|\mathcal{V}_\text{new}|=500$ or $\text{1K}$ when using Llama2 on \textsc{sum} (Figure \ref{fig:target_token_ratio}).
Therefore, although larger $|\mathcal{V}_\text{new}|$ can shorten prompt length in a target language, it does not always guarantee inference speedups unless a model is well-trained to handle $\mathcal{V}_\text{new}$.

\begin{figure}[!t]
\centering
\begin{minipage}{0.49\textwidth}
    \centering
    \includegraphics[width=\textwidth]{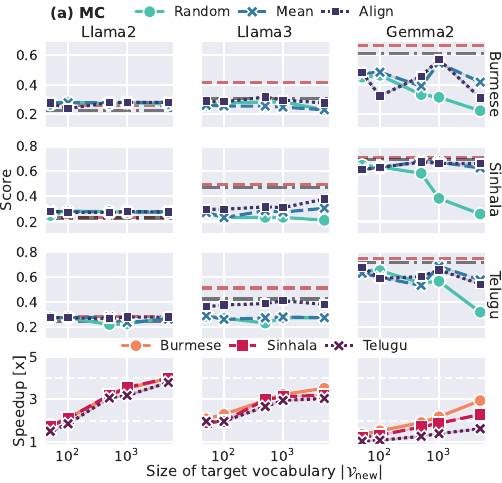}
\end{minipage}
\hfill
\begin{minipage}{0.49\textwidth}
    \centering
    \includegraphics[width=\textwidth]{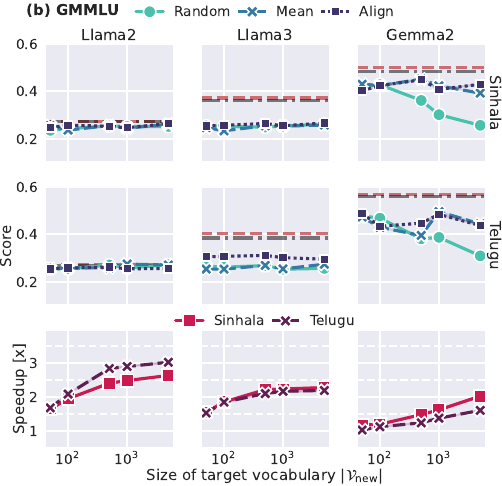}
\end{minipage}
\caption{Downstream performance and inference speedup in (a) \textsc{mc} and (b) \textsc{gmmlu} across different $|\mathcal{V}_\text{new}|$.
\textcolor{red}{Red} and \textcolor{gray}{gray} dotted lines denote Source and CPT-only.
}
\label{fig:performance_by_vocab_cls}
\end{figure}

\subsubsection{Classification Tasks}

Figure \ref{fig:performance_by_vocab_cls} shows the performance changes and corresponding inference speedups with respect to different $|\mathcal{V}_\text{new}|$ on classification tasks.
Unlike generation tasks, larger $|\mathcal{V}_\text{new}|$ does not always result in worse task performance. This is especially evident in Llama2 and Llama3 across tasks.
Nonetheless, Gemma2 adapted models with Random often exhibit performance degradation as $|\mathcal{V}_\text{new}|$ becomes larger across tasks and languages, suggesting the importance of target parameter initialization.

We hypothesize that the observed difference between classification and generation tasks stems from the fundamental nature of the task, as discussed in \S\ref{subsec:performance} and \S\ref{subsec:llama3}. Classification often involves predicting only a single token, which requires less extensive and nuanced generation capabilities compared to generation tasks, where models must produce multi-token sequences.
Therefore, a model might not need to be as well-trained to effectively utilize $\mathcal{V}_\text{new}$ for inference in classification, as it does not generate sequences incorporating those new tokens. 
While performance degradation is substantial in Llama3 and Gemma2 (as observed in \S\ref{subsec:llama3}), classification tasks generally appear more robust to changes in $|\mathcal{V}_\text{new}|$.
This also explains why inference speedups tend not to plateau in most classification scenarios, as larger $|\mathcal{V}_\text{new}|$ can effectively shorten prompt length in a target language (Table \ref{tab:tokenization} for an qualitative example), leading to continued speedup gains without the performance trade-offs seen on generative tasks.

\paragraph{Recommendation}
Setting $|\mathcal{V}_\text{new}|$ between 500 and 1K for \textsc{mt} and 100 for \textsc{sum} can be a suitable starting point to maintain competitive performance to Source while benefiting greatly from inference speedups under low-resource settings. While classification tasks are more robust to changes in $|\mathcal{V}_\text{new}|$, it is often preferable to utilize either CPT-only or Source for better performance if: (i) inference speedups are not the primary priority, and (ii) a post-hoc, training-free performance degradation mitigation technique, as discussed later in \S\ref{subsec:elchat}, is not applied.

\section{Discussion}
\subsection{Source Knowledge Retention}\label{subsec:source_performance}
Previous work~\cite{tejaswi-etal-2024-exploring,mundra-etal-2024-empirical} has reported that vocabulary expansion followed by LAPT can lead to catastrophic forgetting of the original capabilities of a source model. We measure the extent to which adapted models in low-resource settings (30K sentences) suffer from this phenomenon by evaluating them on English-centric reading comprehension and general knowledge and reasoning benchmarks. To this end, we employ the English subset of Belebele (\textsc{mc}) and \textsc{mmlu}~\cite{hendrycks2021measuring}.
We use all three source models and follow the same setup as in \S\ref{subsec:llama3} and \S\ref{subsec:vocab_size}.
However, for brevity, we evaluate only the best-performing Align initialization method. Table \ref{tab:eng_result} presents the corresponding results on \textsc{mc} and \textsc{mmlu}.

\begin{table}[t]
\caption{
Source knowledge retention evaluation on English benchmarks. Each language name indicates that a model has undergone LAPT on its corresponding target language data (30K sentences).
}
\label{tab:eng_result}
\begin{center} %
    \begin{minipage}[t]{0.48\textwidth} %
        \centering
        \small
        \begin{tabular}{lcccccc}
            \toprule
            & \multicolumn{2}{c}{\textbf{Burmese}}
            & \multicolumn{2}{c}{\textbf{Sinhala}}
            & \multicolumn{2}{c}{\textbf{Telugu}}\\
            \textbf{Llama2} & \multicolumn{1}{c}{\scriptsize \textsc{mc}} & \multicolumn{1}{c}{\scriptsize \textsc{mmlu}}
            & \multicolumn{1}{c}{\scriptsize \textsc{mc}} & \multicolumn{1}{c}{\scriptsize \textsc{mmlu}}
            & \multicolumn{1}{c}{\scriptsize \textsc{mc}} & \multicolumn{1}{c}{\scriptsize \textsc{mmlu}}\\
            \midrule
            \rowcolor{gray!25}
            Source & .52 & .46 & .52 & .46 & .52 & .46\\
            CPT-only & .53 & .47 & .52 & .47 & .52 & .46\\
            \midrule
            Align & .38 & .37 & .42 & .39 & .41 & .37\\
            \bottomrule
        \end{tabular}
    \end{minipage}%
    \hfill %
    \begin{minipage}[t]{0.48\textwidth} %
        \centering
        \small
        \begin{tabular}{lcccccc}
            \toprule
            & \multicolumn{2}{c}{\textbf{Burmese}}
            & \multicolumn{2}{c}{\textbf{Sinhala}}
            & \multicolumn{2}{c}{\textbf{Telugu}}\\
            \textbf{Llama3} & \multicolumn{1}{c}{\scriptsize \textsc{mc}} & \multicolumn{1}{c}{\scriptsize \textsc{mmlu}}
            & \multicolumn{1}{c}{\scriptsize \textsc{mc}} & \multicolumn{1}{c}{\scriptsize \textsc{mmlu}}
            & \multicolumn{1}{c}{\scriptsize \textsc{mc}} & \multicolumn{1}{c}{\scriptsize \textsc{mmlu}}\\
            \midrule
            \rowcolor{gray!25}
            Source & .88 & .67 & .88 & .67 & .88 & .67\\
            CPT-only & .86 & .66 & .86 & .66 & .87 & .66\\
            \midrule
            Align & .75 & .46 & .65 & .39 & .78 & .50\\
            \bottomrule
        \end{tabular}
    \end{minipage}

    \vspace{1em} %

    \begin{minipage}{\textwidth} %
        \centering
        \small
        \begin{tabular}{lcccccc}
            \toprule
            & \multicolumn{2}{c}{\textbf{Burmese}}
            & \multicolumn{2}{c}{\textbf{Sinhala}}
            & \multicolumn{2}{c}{\textbf{Telugu}}\\
            \textbf{Gemma2} & \multicolumn{1}{c}{\scriptsize \textsc{mc}} & \multicolumn{1}{c}{\scriptsize \textsc{mmlu}}
            & \multicolumn{1}{c}{\scriptsize \textsc{mc}} & \multicolumn{1}{c}{\scriptsize \textsc{mmlu}}
            & \multicolumn{1}{c}{\scriptsize \textsc{mc}} & \multicolumn{1}{c}{\scriptsize \textsc{mmlu}}\\
            \midrule
            \rowcolor{gray!25}
            Source & .91 & .72 & .91 & .72 & .91 & .72\\
            CPT-only & .90 & .72 & .91 & .72 & .90 & .72\\
            \midrule
            Align & .88 & .67 & .89 & .66 & .84 & .63\\
            \bottomrule
        \end{tabular}
    \end{minipage}
\end{center}
\end{table}

Overall, we observe varying degrees of source knowledge retention depending on the adaptation method. 
CPT-only (i.e. continual per-training without vocabulary expansion) generally preserves original source model capabilities, showing negligible performance drops of up to 2 points across different base models.
In contrast, adapted models initialized with Align exhibit moderate to substantial performance degradation on these English tasks, with average drops ranging from 5.3 points for Gemma2, to 10 points for Llama2, and a substantial 18.7 points for Llama3.
These results corroborate the findings in \citet{mundra-etal-2024-empirical}, which noted that performance degradation in source language tasks tends to be more severe during early LAPT stages with vocabulary expansion.
This suggests that while LAPT with Align is effective for target language adaptation, especially on generation tasks, it comes with a considerable trade-off in the retention of the original source capabilities.

\subsection{Can we recover performance degradation from vocabulary expansion in low-resource settings without training?} \label{subsec:elchat}

As discussed in \S\ref{sec:results}, vocabulary expansion in low-resource settings offers inference speedups and largely retains or even improves performance on generation tasks.
However, it leads to performance degradation in (i) target language classification tasks (\S\ref{subsec:llama3} and \S\ref{subsec:vocab_size}) and (ii) source language capabilities (\S\ref{subsec:source_performance}). 
This section investigates whether such performance degradation can be recovered without additional training.
Indeed, our work, ElChat \cite{yamaguchi2025elchatadaptingchatlanguage}, provides a post-hoc, training-free method to recover the original capabilities of adapted models.

\paragraph{ElChat: A Post-hoc Method to Mitigate Catastrophic Forgetting}
ElChat~\cite{yamaguchi2025elchatadaptingchatlanguage} is a post-hoc, training-free method designed to restore the original capabilities of an LLM after it has undergone vocabulary expansion and continual pre-training on target language data.
The method requires access to an instruction-tuned model that has been further supervised fine-tuned on labeled conversational data, enabling the source base model (i.e. $\mathcal{M}_\text{s}$) to follow instructions.
This prerequisite is readily met in practice, as frontier models like Llama3 and Gemma 2 commonly offer both base and instruction-tuned variants.
ElChat addresses catastrophic forgetting of original capabilities by employing a strategic combination of two core steps: model merging and copying special token weights.
This allows for robust recovery of degraded performance without incurring further training costs.\footnote{For more technical details, please refer to \citet{yamaguchi2025elchatadaptingchatlanguage}.}

\setlength{\tabcolsep}{4pt}
\renewcommand*{\arraystretch}{1.0}
\begin{table}[!t]
\centering
\small
\caption{Gemma2 performance and inference speedup using ElChat, a post-hoc training-free method to mitigate catastrophic forgetting. \colorbox{green!20}{Green} indicates positive performance change over Source.
}

\begin{center}
\resizebox{\linewidth}{!}{%
\begin{tabular}{llllllllllllll}
\multicolumn{14}{c}{(a) Target language tasks}\\
\toprule
 & & \multicolumn{4}{c}{\textbf{Burmese}}  
 & \multicolumn{4}{c}{\textbf{Sinhala}} 
 & \multicolumn{4}{c}{\textbf{Telugu}}\\

\cmidrule(lr){3-6}
\cmidrule(lr){7-10}
\cmidrule(lr){11-14}

& & \multicolumn{1}{c}{\scriptsize \textsc{mt}} & \multicolumn{1}{c}{\scriptsize \textsc{sum}}
& \multicolumn{1}{c}{\scriptsize \textsc{mc}} & \multicolumn{1}{c}{\scriptsize \textsc{gmmlu}}
& \multicolumn{1}{c}{\scriptsize \textsc{mt}} & \multicolumn{1}{c}{\scriptsize \textsc{sum}}
& \multicolumn{1}{c}{\scriptsize \textsc{mc}} & \multicolumn{1}{c}{\scriptsize \textsc{gmmlu}}
& \multicolumn{1}{c}{\scriptsize \textsc{mt}} & \multicolumn{1}{c}{\scriptsize \textsc{sum}}
& \multicolumn{1}{c}{\scriptsize \textsc{mc}} & \multicolumn{1}{c}{\scriptsize \textsc{gmmlu}}\\

\midrule
\rowcolor{gray!25}

& Source & .21\textsubscript{{\tiny.01}} & 34\textsubscript{{\tiny0.1}} & .67 & -
& .19\textsubscript{{\tiny.00}} & 34\textsubscript{{\tiny0.2}} & .71 & .50
& .31\textsubscript{{\tiny.01}} & 30\textsubscript{{\tiny0.2}} & .74 & .57\\

& CPT-only & \cellcolor{green!20}.28\textsubscript{{\tiny.00}} & 28\textsubscript{{\tiny0.3}} & .61 & -
& \cellcolor{green!20}.29\textsubscript{{\tiny.00}} & 34\textsubscript{{\tiny0.1}} & .69 & .48
& \cellcolor{green!20}.43\textsubscript{{\tiny.00}} & 29\textsubscript{{\tiny0.1}} & .71 & .56\\

\midrule
\multirow{2}{*}{\rotatebox[origin=c]{90}{\parbox[c]{0.75cm}{\centering \tiny \textbf{+Speedup}}}} 

& Align & \cellcolor{green!20}.25\textsubscript{{\tiny.00}} & 25\textsubscript{{\tiny0.2}} & .32 & - 
& \cellcolor{green!20}.28\textsubscript{{\tiny.00}} & 32\textsubscript{{\tiny0.1}} & .63 & .42
& \cellcolor{green!20}.32\textsubscript{{\tiny.00}} & 29\textsubscript{{\tiny0.1}} & .59 & .43\\
& + ElChat & \cellcolor{green!20}.30\textsubscript{{\tiny.00}} & 33\textsubscript{{\tiny0.1}} & .64 & -  & \cellcolor{green!20}.31\textsubscript{{\tiny.00}} & \cellcolor{green!20}35\textsubscript{{\tiny0.1}} & \cellcolor{green!20}.73 & .46 & \cellcolor{green!20}.41\textsubscript{{\tiny.00}} & \cellcolor{green!20}31\textsubscript{{\tiny0.1}} & .74 & .52\\

\midrule
\multicolumn{2}{l}{\textbf{Speedup}} & 1.52{\small x} & 1.57{\small x} & 1.51{\small x} & - & 1.26{\small x} & 1.38{\small x} & 1.31{\small x} & 1.20{\small x} & 1.07{\small x} & 1.10{\small x} & 1.06{\small x} & 1.13{\small x} \\
\multicolumn{2}{l}{~~(ElChat)} 
& 1.53{\small x} & 1.73{\small x} & 1.48{\small x} & - & 1.20{\small x} & 1.39{\small x} & 1.30{\small x} & 1.18{\small x} & 1.09{\small x} & 1.14{\small x} & 1.06{\small x} & 1.02{\small x} \\

\bottomrule
\end{tabular}%
}

\vspace{1em}
\begin{tabular}{lcccccc}
            \multicolumn{7}{c}{(b) English tasks}\\
            \toprule
            & \multicolumn{2}{c}{\textbf{Burmese}}
            & \multicolumn{2}{c}{\textbf{Sinhala}}
            & \multicolumn{2}{c}{\textbf{Telugu}}\\
            \textbf{Gemma2} & \multicolumn{1}{c}{\scriptsize \textsc{mc}} & \multicolumn{1}{c}{\scriptsize \textsc{mmlu}}
            & \multicolumn{1}{c}{\scriptsize \textsc{mc}} & \multicolumn{1}{c}{\scriptsize \textsc{mmlu}}
            & \multicolumn{1}{c}{\scriptsize \textsc{mc}} & \multicolumn{1}{c}{\scriptsize \textsc{mmlu}}\\
            \midrule
            \rowcolor{gray!25}
            Source & .91 & .72 & .91 & .72 & .91 & .72\\
            CPT-only & .90 & .72 & .91 & .72 & .90 & .72\\
            \midrule
            Align & .88 & .67 & .89 & .66 & .84 & .63\\
            + ElChat & .93 & .72 & .93 & .71 & .93 & .71\\
            \bottomrule
        \end{tabular}
\label{tab:elchat_result}
\end{center}
\end{table}

\paragraph{Results}
Constrained by resources, we apply ElChat to adapted Gemma2 models initialized with Align to evaluate its impact on both target language and English tasks.\footnote{Gemma2 is chosen for its notable performance degradation in target language tasks (not only in classification tasks but also in Burmese \textsc{sum}) and as a representative of more recent models than Llama2 and Llama3, enabling a practical and meaningful comparison within our budget.}
We observe from Table \ref{tab:elchat_result} (a) that ElChat consistently improves performance across target language tasks. 
Specifically, for classification tasks, the performance drop relative to Source is substantially reduced within 5 points (Telugu \textsc{gmmlu}), a notable improvement from up to 35 points (Burmese \textsc{mc}) without ElChat.
Furthermore, ElChat not only recovers degraded performance but also maintains the superior generative capabilities of the adapted models, even showing improvements in some cases (e.g. Sinhala \textsc{sum}).
Crucially, these performance recoveries are achieved while largely retaining the inference speedups gained from vocabulary expansion.
The effect of ElChat is even more pronounced in English tasks (Table \ref{tab:elchat_result} (b)).
The adapted models almost fully recover their original source capabilities, matching Source performance with a negligible drop of at most 1 point. This highlights the effectiveness of using a post-hoc, training-free method to mitigate performance degradation from vocabulary expansion without requiring any additional training.

\paragraph{Recommendation}
Using a post-hoc, training-free method like ElChat enables an adapted model to restore the original capabilities of the corresponding source model without additional training. For target language tasks, this can lead to enhanced performance while preserving the inference speedups gained from vocabulary expansion.

\subsection{Comparison with Vocabulary Replacement}
Intuitively, vocabulary replacement, which typically replaces the entire source vocabulary with a new one from a target language, necessitates a greater number of training tokens than vocabulary expansion. This is attributed to its substantially larger number of new parameters requiring alignment. For instance, \citet{pmlr-v235-dagan24a} found that over 50 billion tokens were necessary to swap a tokenizer at no performance cost in their domain adaptation experiments. 
To illustrate the challenges of vocabulary replacement for target language adaptation in low-resource settings compared to vocabulary expansion, we conduct a brief comparative study using Gemma2 as the source model.

\paragraph{Experimental Setup}
We first train a target tokenizer with a vocabulary size of 32K using target language data $\mathcal{D}$ and the same training configurations as the base Gemma2 tokenizer.
For target parameter initialization, we consider Random, FOCUS, and Mean. We then continually pre-train each initialized model using the same approach as in Sections \ref{subsec:llama3} and \ref{subsec:vocab_size} (i.e. 2x2 LS+\textsc{mtp}+512).

\setlength{\tabcolsep}{4pt}
\begin{table}[!t]
\centering
\small
\caption{Vocabulary replacement performance and inference speedup on target language tasks using Gemma2 as source. \colorbox{green!20}{Green} indicates positive performance change over Source. The speedup ratio corresponds to Mean.
}

\begin{center}
\resizebox{\linewidth}{!}{%
\begin{tabular}{llllllllllllll}
\toprule
 & & \multicolumn{4}{c}{\textbf{Burmese}}  
 & \multicolumn{4}{c}{\textbf{Sinhala}} 
 & \multicolumn{4}{c}{\textbf{Telugu}}\\

\cmidrule(lr){3-6}
\cmidrule(lr){7-10}
\cmidrule(lr){11-14}

 & & \multicolumn{1}{c}{\scriptsize \textsc{mt}} & \multicolumn{1}{c}{\scriptsize \textsc{sum}}
& \multicolumn{1}{c}{\scriptsize \textsc{mc}} & \multicolumn{1}{c}{\scriptsize \textsc{gmmlu}}
& \multicolumn{1}{c}{\scriptsize \textsc{mt}} & \multicolumn{1}{c}{\scriptsize \textsc{sum}}
& \multicolumn{1}{c}{\scriptsize \textsc{{mc}}} & \multicolumn{1}{c}{\scriptsize \textsc{gmmlu}}
& \multicolumn{1}{c}{\scriptsize \textsc{mt}} & \multicolumn{1}{c}{\scriptsize \textsc{sum}}
& \multicolumn{1}{c}{\scriptsize \textsc{mc}} & \multicolumn{1}{c}{\scriptsize \textsc{gmmlu}}\\

\midrule
\rowcolor{gray!25}
& Source & .21\textsubscript{{\tiny.01}} & 34\textsubscript{{\tiny0.1}} & .67 & -
& .19\textsubscript{{\tiny.00}} & 34\textsubscript{{\tiny0.2}} & .71 & .50
& .31\textsubscript{{\tiny.01}} & 30\textsubscript{{\tiny0.2}} & .74 & .57\\

& CPT-only & \cellcolor{green!20}.28\textsubscript{{\tiny.00}} & 28\textsubscript{{\tiny0.3}} & .61 & -
& \cellcolor{green!20}.29\textsubscript{{\tiny.00}} & 34\textsubscript{{\tiny0.1}} & .69 & .48
& \cellcolor{green!20}.43\textsubscript{{\tiny.00}} & 29\textsubscript{{\tiny0.1}} & .71 & .56\\

\midrule
\multirow{3}{*}{\rotatebox[origin=c]{90}{\parbox[c]{1cm}{\centering \tiny \textbf{+Speedup}}}}
& Random & .00\textsubscript{{\tiny.00}} & 0\textsubscript{{\tiny0.0}} & .22 & -  & .00\textsubscript{{\tiny.00}} & 0\textsubscript{{\tiny0.0}} & .27 & .25 & .00\textsubscript{{\tiny.00}} & 1\textsubscript{{\tiny0.1}} & .24 & .23\\
& FOCUS & .00\textsubscript{{\tiny.00}} & 4\textsubscript{{\tiny0.1}} & .22 & -  & .02\textsubscript{{\tiny.00}} & 12\textsubscript{{\tiny0.3}} & .27 & .27 & .00\textsubscript{{\tiny.00}} & 3\textsubscript{{\tiny0.0}} & .23 & .27\\
& Mean & .06\textsubscript{{\tiny.00}} & 16\textsubscript{{\tiny0.1}} & .26 & -  & .01\textsubscript{{\tiny.00}} & 23\textsubscript{{\tiny0.3}} & .27 & .26 & .04\textsubscript{{\tiny.00}} & 12\textsubscript{{\tiny0.3}} & .34 & .26\\

\midrule
\multicolumn{2}{l}{\textbf{Speedup}} & 1.14{\small x} & 2.27{\small x} & 2.95{\small x} & - & 1.01{\small x} & 1.36{\small x} & 2.70{\small x} & 2.44{\small x} & 0.89{\small x} & 0.57{\small x} & 1.77{\small x} & 1.91{\small x} \\
\bottomrule
\end{tabular}%
}
\label{tab:vr_result}
\end{center}
\end{table}

\paragraph{Results}
Table \ref{tab:vr_result} presents the performance and inference speedups of models adapted using vocabulary replacement.
Overall, these models exhibit notably poor performance across all initialization approaches and tasks in low-resource settings. Specifically, Random and FOCUS largely fail to perform well on generation tasks, yielding near-zero scores in \textsc{mt} and low scores of up to 12 points in \textsc{sum}.
While Mean shows slight improvement over Random and FOCUS, its performance remains substantially lower than that of Source, CPT-only, and crucially, the vocabulary expansion counterpart (Table \ref{tab:llama3_result}).
For instance, in Telugu \textsc{mc}, the best vocabulary replacement performance (Mean at 34 points) is still far below Source (74) and vocabulary expansion with Mean (60).
The only advantage of vocabulary replacement lies in its superior inference speedups in some tasks (e.g. Burmese \textsc{mc} at 2.95x, Sinhala \textsc{mc} at 2.70x, and Telugu \textsc{gmmlu} at 1.91x), which stems from its entirely optimized vocabulary for the target language. 
These results clearly demonstrate the significant challenges of effectively applying vocabulary replacement for target language adaptation under low-resource settings, when compared to vocabulary expansion.

\subsection{Limitations}
\paragraph{Target Language Classification Performance}
While we demonstrate that ElChat, a post-hoc and training-free method, helps adapted models recover degraded performance in target language classification tasks, a moderate performance drop can still persist, as observed in Telugu \textsc{gmmlu} (Table \ref{tab:elchat_result}). 
Furthermore, ElChat is not applicable in scenarios where only base models are available as source for adaptation (i.e. lacking an instruction-tuned variant).
These limitations represent ongoing challenges that extend beyond the immediate scope of this article.

\paragraph{Comparison with Fully Multilingual LLMs}
Due to resource constraints, this article does not include a direct comparison with fully multilingual models like MaLA-500~\cite{Lin2024MaLA500ML} and EMMA-500~\cite{ji2025emma500enhancingmassivelymultilingual,ji2025massivelymultilingualadaptationlarge}. Such a comparison would help contextualize our results.

\paragraph{Model Size}
Our experiments use models of up to 9B parameters due to resource constraints. Scaling these experiments to larger LLMs is a valuable direction for future work to confirm the generalizability of our findings.

\paragraph{Tokenizer}
Heuristic-based target parameter initialization with Merge assumes the use of a BPE-based tokenizer which is a common choice in recent LLMs, e.g. Gemma2, Llama3, Llama2, Mistral~\cite{Jiang2023Mistral7}, \textit{inter alia}.
Experimenting with other tokenizers, such as Unigram~\cite{kudo-2018-subword}, falls outside the scope of this article.

\section{Conclusion}
We investigated cross-lingual vocabulary expansion in low-resource settings across target parameter initialization approaches and training strategies.
Our extensive experiments reveal that a widely used approach in high-resource settings is not always optimal in low-resource settings.
In contrast, models adapted by our alternative strategies achieve faster inference while rivaling their base models in task performance, especially on generation tasks.
We supplement our analysis with specific recommendations for effective vocabulary expansion.

\appendix
\appendixsection{Details on Experimental Setup} \label{appendix:setup}

Table \ref{tab:prompt} lists prompt templates for all tasks, while Table \ref{tab:hyperparams_pretraining} lists the hyperparameters used for both (a) LAPT and (b) inference.

\begin{table}[t]
\begin{center}
\caption{
Prompt template for each task and language. For \textsc{mc} and \textsc{gmmlu}, we omit text breaks (\textbackslash n) after context, question, and each option for readability.
}
\includegraphics[width=\columnwidth]{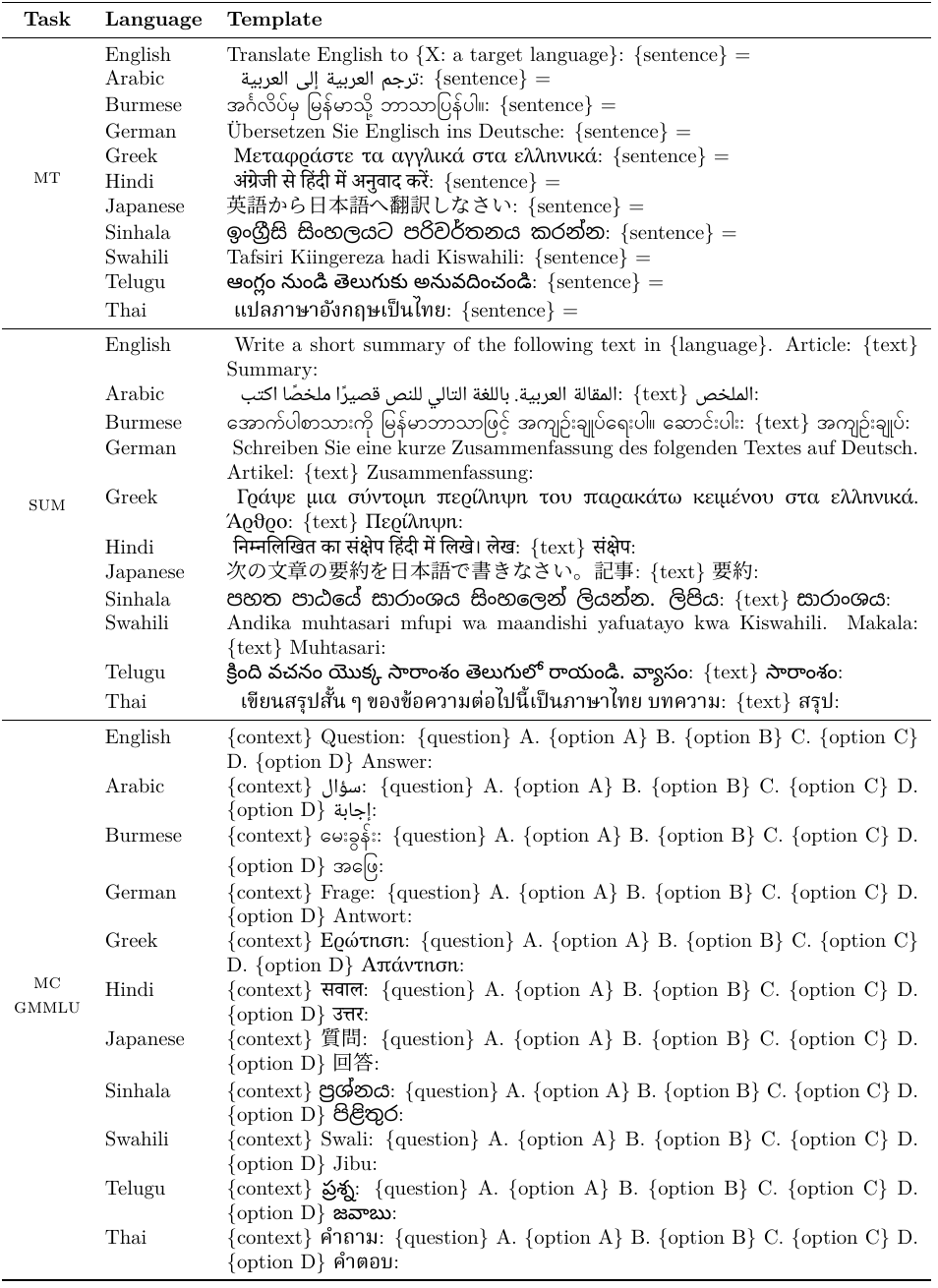}
\label{tab:prompt}
\end{center}
\end{table}

\begin{table}[t]
\centering
\small
\caption{Hyperparameter configurations for LAPT and inference.}
\begin{minipage}{0.48\textwidth}
    \centering
    \begin{tabular}{lc}
        \multicolumn{2}{c}{(a) LAPT}\\
        \toprule
        \textbf{Hyperparameters} & \textbf{Values} \\
        \midrule
        Batch size & 8\\
        Maximum number of training epochs & 2\\
        Adam $\epsilon$ & 1e-8\\
        Adam $\beta_1$ & 0.9\\
        Adam $\beta_2$ & 0.999\\
        Sequence length & 2,048\\
        Learning rate & 1e-4\\
        Learning rate scheduler & cosine\\
        Warmup steps & 100 \\
        Weight decay & 0.01\\
        Attention dropout & 0.0 \\
        Dropout & 0.05\\
        LoRA rank $r$ & 8\\
        LoRA dropout & 0.05\\
        LoRA $\alpha$ & 32 \\
        \bottomrule
    \end{tabular}
    \label{tab:hyperparams_pretraining}
\end{minipage}
\hfill
\begin{minipage}{0.48\textwidth}
    \centering
    \resizebox{\textwidth}{!}{%
        \begin{tabular}{lc}
            \multicolumn{2}{c}{(b) Inference}\\
            \toprule
            \textbf{Parameters} & \textbf{Values} \\
            \midrule
            Maximum prompt length & 4,096\\
            Temperature & 0.8\\
            Repetition penalty & 1.1\\
            Top $k$ & 40\\
            Top $p$ & 0.9\\
            Beam width & 5\\
            Sampling & True\\
            Early stopping & True\\
            Maximum number of generated tokens & 128\\
            \bottomrule
        \end{tabular}%
    }
    \label{tab:params_generation}
\end{minipage}
\end{table}

\appendixsection{Supplementary Results} \label{appendix:results}

Table \ref{tab:lapt_strategy_cls} lists the performance of the adapted models with different training strategies on classification tasks.

Table \ref{tab:tokenization} visualizes how tokenization changes with vocabulary expansion in Burmese using Llama2 as source.

\setlength{\tabcolsep}{3pt}
\begin{table}[!t]

\caption{
Mean performance of Align models on classification tasks with Llama2 as source.
\colorbox{green!20}{Green} indicates positive performance change over Source.
}

\begin{center}
\small
\begin{tabular}{llllllllllllll}
\toprule
 & & \multicolumn{2}{c}{\textbf{Arabic}}  
 & \multicolumn{2}{c}{\textbf{Burmese}}
 & \multicolumn{2}{c}{\textbf{Greek}} 
 & \multicolumn{2}{c}{\textbf{Hindi}}
 & \multicolumn{2}{c}{\textbf{Sinhala}}
 & \multicolumn{2}{c}{\textbf{Telugu}}\\

\cmidrule(lr){3-4}
\cmidrule(lr){5-6}
\cmidrule(lr){7-8}
\cmidrule(lr){9-10}
\cmidrule(lr){11-12}
\cmidrule(lr){13-14}

\multicolumn{2}{c}{\textbf{Model}} & \multicolumn{1}{c}{\scriptsize \textsc{mc}} & \multicolumn{1}{c}{\scriptsize \textsc{gmmlu}}
& \multicolumn{1}{c}{\scriptsize \textsc{mc}} & \multicolumn{1}{c}{\scriptsize \textsc{gmmlu}}
& \multicolumn{1}{c}{\scriptsize \textsc{mc}} & \multicolumn{1}{c}{\scriptsize \textsc{gmmlu}}
& \multicolumn{1}{c}{\scriptsize \textsc{mc}} & \multicolumn{1}{c}{\scriptsize \textsc{gmmlu}}
& \multicolumn{1}{c}{\scriptsize \textsc{mc}} & \multicolumn{1}{c}{\scriptsize \textsc{gmmlu}}
& \multicolumn{1}{c}{\scriptsize \textsc{mc}} & \multicolumn{1}{c}{\scriptsize \textsc{gmmlu}}\\
\midrule

\rowcolor{gray!25}
& Source & .29 & .29 & .26 & -  & .27 & .28 & .25 & .28 & .24 & .27 & .28 & .27\\

& CPT-only & \cellcolor{green!20}.30 & .29 & .22 & -  & \cellcolor{green!20}.29 & \cellcolor{green!20}.29 & \cellcolor{green!20}.28 & .28 & .23 & .27 & .24 & .27\\
\midrule
\multirow{4}{*}{\rotatebox[origin=c]{90}{\parbox[c]{1.25cm}{\centering \footnotesize \textbf{LoRA}}}}
& \textsc{clm}+2048 & .28 & .25 & .26 & -  & \cellcolor{green!20}.31 & .26 & \cellcolor{green!20}.27 & .27 & \cellcolor{green!20}.27 & .27 & .22 & .27\\
& \textsc{mtp}+2048 & .29 & .26 & \cellcolor{green!20}.28 & -  & .26 & .28 & \cellcolor{green!20}.28 & .25 & \cellcolor{green!20}.28 & .26 & .27 & .27\\
& \textsc{clm}+512 & .28 & .25 & .24 & -  & \cellcolor{green!20}.33 & .26 & \cellcolor{green!20}.28 & .27 & \cellcolor{green!20}.27 & .26 & .27 & .26\\
& \textsc{mtp}+512 & .29 & .28 & .24 & -  & .27 & .27 & \cellcolor{green!20}.30 & .26 & .24 & .25 & .27 & .27\\
\midrule
\multirow{4}{*}{\rotatebox[origin=c]{90}{\parbox[c]{1.25cm}{\centering \footnotesize \textbf{2-stage}}}}
& \textsc{clm}+2048 & \cellcolor{green!20}.30 & .27 & .24 & -  & .26 & .25 & \cellcolor{green!20}.27 & .26 & \cellcolor{green!20}.27 & .27 & .27 & .26\\
& \textsc{mtp}+2048 & .29 & .27 & .22 & -  & .23 & .24 & \cellcolor{green!20}.28 & .26 & \cellcolor{green!20}.29 & .26 & .27 & .26\\
& \textsc{clm}+512 & .25 & .28 & .22 & -  & \cellcolor{green!20}.29 & .27 & \cellcolor{green!20}.27 & .27 & \cellcolor{green!20}.27 & .25 & .26 & .27\\
& \textsc{mtp}+512 & .23 & .28 & .22 & -  & .27 & .28 & \cellcolor{green!20}.29 & .26 & \cellcolor{green!20}.28 & .25 & .22 & .27\\
\midrule
\multirow{4}{*}{\rotatebox[origin=c]{90}{\parbox[c]{1.25cm}{\centering \footnotesize \textbf{2x2 LS}}}}
& \textsc{clm}+2048 & .29 & .27 & .22 & -  & \cellcolor{green!20}.29 & .24 & \cellcolor{green!20}.28 & .26 & \cellcolor{green!20}.27 & .26 & .28 & .26\\
& \textsc{mtp}+2048 & \cellcolor{green!20}.30 & .27 & .26 & -  & \cellcolor{green!20}.32 & .25 & \cellcolor{green!20}.28 & .27 & \cellcolor{green!20}.28 & .24 & \cellcolor{green!20}.29 & .27\\
& \textsc{clm}+512 & \cellcolor{green!20}.31 & .29 & .22 & -  & \cellcolor{green!20}.28 & .24 & \cellcolor{green!20}.28 & .28 & \cellcolor{green!20}.27 & .26 & .28 & .26\\
& \textsc{mtp}+512 & .28 & .27 & .24 & -  & .23 & .23 & \cellcolor{green!20}.28 & .26 & \cellcolor{green!20}.27 & .26 & .27 & .26\\
\bottomrule
\end{tabular}%
\label{tab:lapt_strategy_cls}
\end{center}
\end{table}

\begin{table}[t]
\begin{center}
\caption{
Example of tokenization changes with vocabulary expansion in Burmese using Llama2 as source. `\textunderscore' stands for a whitespace.
}
\includegraphics[width=\columnwidth]{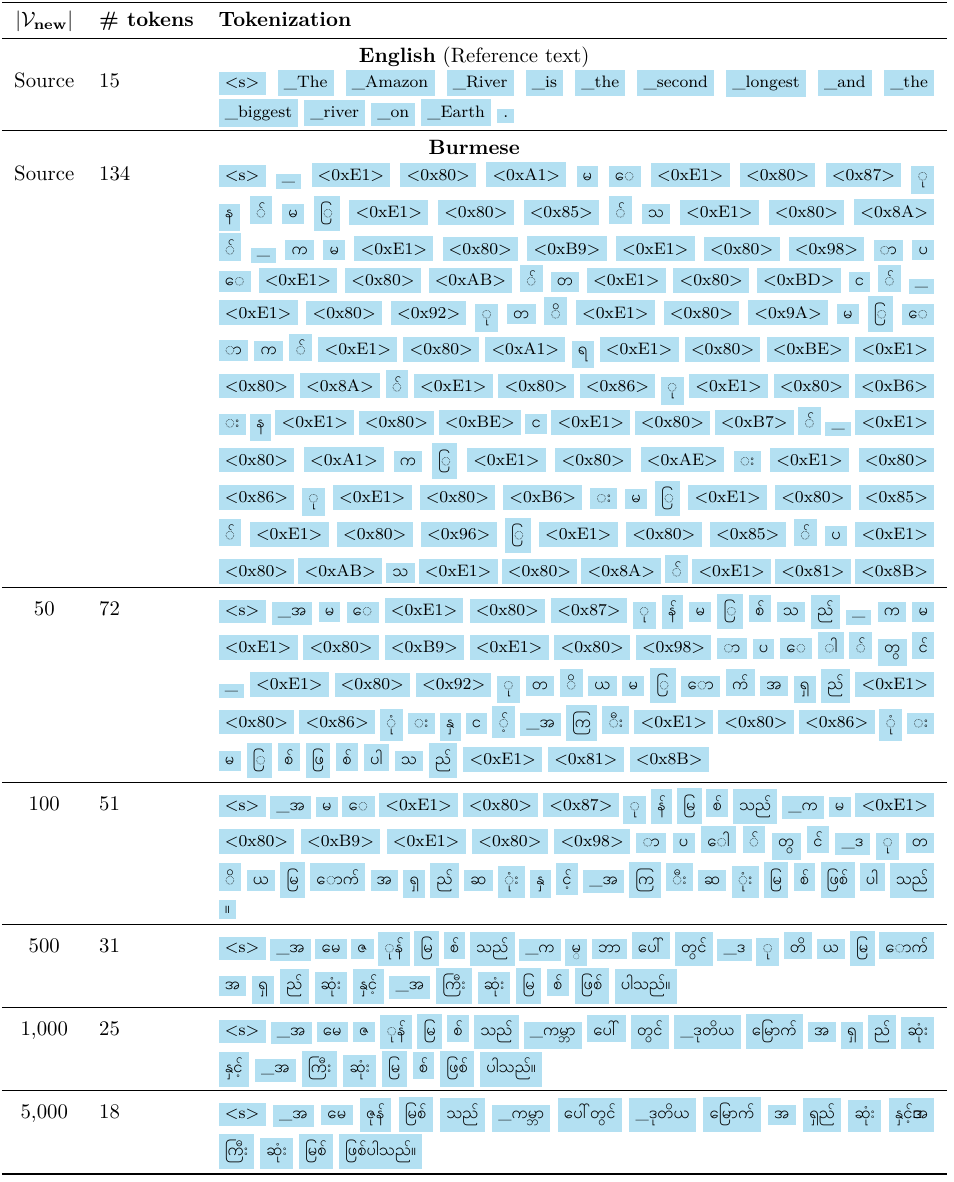}
\label{tab:tokenization}
\end{center}
\end{table}

\begin{acknowledgments}
We thank the anonymous reviewers and Action Editor, Minlie Huang, for their constructive and detailed feedback.
We are also grateful to Miles Williams, Huiyin Xue, and Constantinos Karouzos for their valuable feedback on the initial draft.
We acknowledge (1) IT Services at the University of Sheffield for providing high-performance computing services, (2) the EuroHPC Joint Undertaking for awarding us access to MeluXina at LuxProvide, Luxembourg, and (3) the use of time on Tier 2 HPC facility JADE2, funded by the Engineering and Physical Sciences Research Council (EPSRC) (EP/T022205/1).
AY is supported by EPSRC [grant number EP/W524360/1] and the Japan Student Services Organization (JASSO) Student Exchange Support Program (Graduate Scholarship for Degree Seeking Students).
AV research is partly supported by MRC-FAPESP [AIM-Health] and CNPq [406926/2025-5].
\end{acknowledgments}

\newpage
\starttwocolumn
\bibliographystyle{compling}
\bibliography{custom,anthology}

\end{document}